# Ocular-Induced Abnormal Head Posture: Diagnosis and Missing Data Imputation

Saja Al-Dabet[1]*, Sherzod Turaev[1], Nazar Zaki[1], Arif O. Khan[2,3], Luai Eldweik[2,3]

[1] College of Information Technology, United Arab Emirates University, Al Ain, United Arab Emirates
[2] Eye Institute, Cleveland Clinic Abu Dhabi, Abu Dhabi, United Arab Emirates
[3] Cleveland Clinic Lerner College of Medicine of Case Western Reserve University, Cleveland, Ohio, USA

700039885@uaeu.ac.ae, sherzod@uaeu.ac.ae , nzaki@uaeu.ac.ae, arif.khan@mssm.edu, eldweil@clevelandclinicabudhabi.ae

## Abstract

Ocular-induced abnormal head posture (AHP) is a compensatory mechanism that arises from ocular misalignment conditions, such as strabismus, enabling patients to reduce diplopia and preserve binocular vision. Early diagnosis minimizes morbidity and secondary complications such as facial asymmetry; however, current clinical assessments remain largely subjective and are further complicated by incomplete medical records. This study addresses both challenges through two complementary deep learning frameworks. First, AHP-CADNet is a multi-level attention fusion framework for automated diagnosis that integrates ocular landmarks, head pose features, and structured clinical attributes to generate interpretable predictions. Second, a curriculum learning–based imputation framework is designed to mitigate missing data by progressively leveraging structured variables and unstructured clinical notes to enhance diagnostic robustness under realistic data conditions. Evaluation on the PoseGaze-AHP dataset demonstrates robust diagnostic performance. AHP-CADNet achieves 96.9%–99.0% accuracy across classification tasks and low prediction errors for continuous variables, with MAE ranging from 0.103 to 0.199 and $R^2$ exceeding 0.93. The imputation framework maintains high accuracy across all clinical variables (93.46%–99.78% with PubMedBERT), with clinical dependency modeling yielding significant improvements ($p < 0.001$). These findings confirm the effectiveness of both frameworks for automated diagnosis and recovery from missing data in clinical settings.

# 1. Introduction

Abnormal head posture (AHP) is a clinical condition characterized by persistent deviation of the head from the neutral position [1]. It is considered a visible postural adaptation that reflects compensatory mechanisms triggered by ocular, neurological, or skeletal causes [2], [3], [4]. Among these causes, ocular-induced AHP holds particular clinical significance, as it typically arises from incomitant strabismus such as Duane syndrome, superior oblique palsy, or Brown syndrome. Patients with ocular misalignment often adopt head postures such as head turn, tilt, chin-up, or chin-down to reduce diplopia, preserve binocular vision, and maintain alignment of the visual axes [4], [5]. Null-point nystagmus is another ocular reason for AHP. Early diagnosis and treatment of ocular-induced AHP minimizes morbidity and reduces the risk of secondary complications. For instance, patients with untreated AHP exhibit a higher prevalence of facial asymmetry compared to those without postural deviations [6], [7]. Long-standing AHPs can also lead to secondary musculoskeletal complications, such as cervical muscle strain, neck pain, and spinal malalignment. Therefore, accurate identification of ocular-induced AHP is crucial, as it not only guides targeted clinical interventions but also serves as an indicator of disease severity.

In clinical practice, patients' data, including those with AHP, are captured in electronic health records (EHRs) as longitudinal data collected throughout the medical care and stored across multiple formats, including structured variables, diagnostic labels, and unstructured clinical notes [8]. Despite their essential role in healthcare systems, EHRs are often affected by data quality deficiencies such as incompleteness, inaccuracy, and lack of plausibility, which can limit their reliability in certain clinical and analytical contexts [9]. Among these issues, missing data is particularly common and typically falls into one of three categories: (a) Missing Completely at Random (MCAR), (b) Missing at Random (MAR), and (c) Missing Not at Random (MNAR) [10]. The most straightforward solution is to exclude incomplete records from further analysis; however, this could lead to other complications, such as a reduced sample size, alongside the potential for introducing bias.

To mitigate the impact of missing data, various imputation strategies have been developed. Methods ranging from statistical models [11], [12], [13] to machine learning (ML) approaches, both supervised and unsupervised [14], [15], have been explored. More recently, deep learning (DL) techniques such as generative adversarial networks (GANs), denoising autoencoders (DAEs), and temporal models like Bidirectional Recurrent Imputation for Time Series (BRITS) have been introduced due to their capability to generalize across diverse missingness patterns [16], [17], [18]. Nevertheless, EHR data remain a valuable resource across a wide range of healthcare applications, including diagnostic systems, clinical decision support, patient monitoring, administrative processes, and population health management [19]. To leverage EHR data effectively, recent studies have emphasized the importance of leveraging both structured and unstructured EHR data, as clinical notes often contain contextual information that complements structured records and enables more comprehensive clinical insight [20], [21], [22]. Similar observations have been made in ophthalmology, where incorporating both structured variables and unstructured clinical data has improved the accuracy of disease identification, while more recent reviews, however, have highlighted persistent gaps in EHR-based ophthalmology studies, particularly in handling missing data [23], [24].

In the context of ocular-induced AHP, these data challenges are particularly relevant. Although the clinical importance of this condition has been well established, the integration of automated diagnostic tools into a real-world context remains limited. Moreover, the issue of missing data has not been addressed in this domain. To investigate these challenges, the previously published PoseGaze-AHP dataset [25] is utilized. This dataset includes 3D image data capturing synchronized head pose and gaze information, along with structured clinical attributes such as reported symptoms and diagnostic labels specific to ocular-induced AHP. The dataset was constructed based on systematically extracted clinical information from peer-reviewed medical research papers [26]. The structured clinical attributes in the dataset are representative of information typically recorded in EHR, enabling their use in EHR-based diagnostic and imputation tasks.

In this paper, two DL–based frameworks are proposed, both developed to improve diagnostic accuracy and data completeness for ocular-induced AHP. The first, AHP-CADNet, is introduced as a multi-level attention fusion framework for automated diagnosis, integrating ocular landmarks, head pose features, and structured clinical variables. The second, a curriculum learning–based imputation model, is proposed to impute missing data in structured attributes from unstructured clinical notes, thereby enhancing diagnostic reliability in the presence of incomplete records. The contributions of this work can be summarized as follows:

- Introduce AHP-CADNet: A multi-level attention fusion framework is proposed, which integrates ocular landmarks, head pose features, and structured clinical attributes. The model employs multi-level attention mechanisms to capture both intra- and inter-modal relationships, utilizing a gated relevance mechanism to enhance diagnostic performance.
- Introduce Curriculum Learning–Based Imputation: A progressive DL framework is proposed for imputing missing data, leveraging both structured variables and clinical notes. This model is designed to handle increasingly complex missingness patterns while preserving clinical relevance.

The remainder of this paper is organized as follows. Section 2 reviews related work on diagnostic and data imputation frameworks. Section 3 introduces the proposed AHP-CADNet and the curriculum learning–based imputation frameworks. Section 4 describes the PoseGaze-AHP dataset, while Section 5 details the experimental settings and Section 6 outlines the evaluation measures. Section 7 presents the results and their analysis. Finally, Section 8 concludes the paper and discusses future research directions.

## 2. Literature Review

The diagnosis of ocular-induced AHP requires addressing two fundamental challenges. The first is accurate detection of both the underlying ocular misalignment, such as strabismus, superior oblique palsy, or Duane syndrome and the compensatory AHPs, including head turn, tilt, chin-up, or chin-down, that patients adopt to preserve binocular vision. Existing methods typically address these aspects separately; head pose estimation ignores ocular causes, while strabismus detection overlooks postural consequences. The second challenge arises from the reliance of clinical AHP assessment on multiple documentation techniques for clinical data that could be incomplete in the EHR. Current diagnostic frameworks do not adequately address such missingness, which limits their clinical applicability. This section reviews these parallel research streams and highlights the gaps that necessitate complementary approaches to advance diagnostic accuracy and address incomplete clinical data.

*2.1 Detecting Abnormal Head Posture and Ocular Misalignment using DL*

The automated analysis of ocular-induced AHP requires consideration of both the underlying ocular pathology and its compensatory postural manifestations. Research in this domain has developed along two parallel tracks, generic head pose estimation and ocular misalignment detection, each with distinct limitations for clinical diagnosis. Current head pose estimation methods achieve impressive technical performance but fail to address the clinical requirements of AHP diagnosis. Liu et al. (2021) introduced a label-free framework that reconstructs personalized 3D face models from single RGB images, achieving MAEs between 4.78° and 7.05° across benchmark datasets. However, their iterative optimization procedure was computationally demanding and sensitive to occlusion, which limited its clinical applicability [27]. Chen et al. (2023) advanced this approach by fusing RGB and depth data through multimodal self-attention networks, achieving state-of-the-art MAEs of 0.84° and 0.93°. Despite superior accuracy, the requirement for depth sensors restricts deployment in typical clinical environments [28].

Structural approaches have shown promise for postural analysis but lack clinical specificity. Lee et al. (2024) developed a framework for forward head posture detection using 2D keypoints and graph convolutional networks, achieving an accuracy of 78.27% across 2,387 samples validated by physical therapists. However, the coarse binary categorization cannot distinguish the subtle compensatory positions characteristic of ocular-induced AHP, such as slight head tilts or chin adjustments [29]. Transformer-based methods have improved robustness under challenging conditions. Dhingra et al. (2022) proposed HeadPosr, which integrates a CNN backbone with transformer encoders to regress yaw, pitch, and roll from single images, achieving MAEs of 5.26° (AFLW2000) and 3.71° (BIWI) [30], while Liu et al. (2023) developed TokenHPE, a token-driven transformer model that represents intra- and cross-orientation facial relationships through visual and orientation tokens. This framework improved performance under occlusion (MAEs of 4.22° and 2.95°) [31]. These advances demonstrate technical progress in generic head pose estimation but do not distinguish clinically abnormal postures or address their ocular origins.

Parallel research has focused on ocular misalignment, a cause of AHP, although it is typically studied in isolation. Chen et al. (2018a) demonstrated the feasibility of objective strabismus diagnosis using eye-tracking–based systems, where handcrafted fixation deviation features supported the detection of strabismus type, affected eye, and severity consistent with clinical evaluation [32]. Building on this, Chen et al. (2018b) introduced a DL approach, converting fixation sequences into gaze deviation images that CNNs processed. With VGG-S, the system achieved 95.2% accuracy, 94.1% sensitivity, and 96.0% specificity, marking a clear advance over handcrafted methods [33]. More accessible solutions have since emerged through mobile platforms. Mesquita et al. (2021) implemented a corneal light reflex–based application for pediatric strabismus screening, achieving accuracies of 84.5% at a 6 PD cutoff and 92.8% at an 11 PD cutoff [34]. Similarly, Huang et al. (2021) applied an image-processing pipeline to facial photographs, utilizing landmark detection and circle fitting to quantify asymmetry in pupil–canthus distances, with a statistically significant separation between strabismic and normal cases ($p < 0.001$) [35]. Recently, Wang et al. (2024) integrated the alternating cover test with a mobile deep learning system, combining Faster-RCNN, Efficient-UNet, and displacement analysis across 2000 images and 109 videos. The framework achieved an AUC of 0.901, with a sensitivity of 96.91% and a specificity of 83.33% [36].

These parallel research tracks expose a fundamental limitation: existing approaches treat ocular pathology and postural compensation as independent problems. Head pose methods concentrate on generic orientation estimation without accounting for ocular causes, measuring broad movements such as yaw, pitch, and roll that cannot distinguish compensatory AHP from normal head motion. In contrast, strabismus detection systems achieve high accuracy in identifying misalignment but overlook the postural adaptations patients develop to manage diplopia and preserve binocular vision. This fragmented perspective fails to capture the clinical reality that AHP arises directly from ocular misalignment, with the severity and direction of head positioning closely correlated to the type and magnitude of the underlying ocular condition. Both research streams also rely heavily on controlled datasets that do not reflect clinical variability. Head pose methods are often trained on small, non-clinical datasets [27] or require specialized hardware unsuitable for routine practice [28]. Strabismus detection research, while clinically motivated, tends to focus on binary classification as present or absent rather than the detailed characterization of misalignment type, affected eye, and severity that is necessary for comprehensive AHP assessment. To overcome these limitations, this study introduces AHP-CADNet, a multi-level attention fusion framework that integrates ocular landmarks, head pose features, and structured clinical attributes. Unlike fragmented prior approaches, the proposed framework explicitly models the relationship between ocular misalignment and compensatory head positioning, producing clinically interpretable predictions for both domains simultaneously. Table 1 summarizes the recent studies in head pose estimation.

Table 1: Summary of Recent Head Pose Estimation and Strabismus Detection

| Study | Data | Method | Target | Results |
|---|---|---|---|---|
| Liu et al. (2021) | Pointing'04, BIWI, AFLW2000, Multi-PIE, Pandora | CNN + 2D–3D alignment | Head Pose | MAE*: 4.78° (Pointing'04), 6.83° (BIWI), 7.05° (AFLW2000) |
| Chen et al. (2023) | BIWI RGB-D | ResNet + PointNet fusion | | MAE: 0.84° (Gumbel), 0.93° (Gaussian) |
| Lee et al. (2024) | StateFarm + public pose datasets (2,387 annotated) | Detectron2 + GCN | | Acc*: 78.3%, F1*: 77.5% |
| Wang et al. (2024) | Watch-n-Patch, BIWI, Pandora, ICT-3DHP | Depth-CNN + clustering | | MAE: 2.74° (BIWI), 3.12° (Pandora) |
| Dhingra et al. (2022) | 300W-LP ; AFLW2000, BIWI | CNN + Transformer | | MAE: 5.26 (AFLW2000), 3.71 (BIWI) |
| Liu et al. (2023) | AFLW2000, BIWI | Token-based Transformer | | MAE: 4.22 (AFLW2000), 2.95° (BIWI) |
| Chen et al. (2018a) | Eye-tracking points | Rule-based + handcrafted features | Strabismus | - |
| Chen et al. (2018b) | Eye-tracking points | CNNs + SVM | | Acc: 95.2%, Sens: 94.1%, Spec: 96.0% |
| Mesquita et al. (2021) | Smartphone images | Rule-based image analysis | | Acc: 84.5%, Sens: 89.5% |
| Huang et al. (2021) | Facial Images | CNN + landmark + binarization | | p < 0.001 |

| Wang et al. (2024) | Images, Videos | Mobile DL + cover test | AUC: 0.901; Sens: 96.9%; Spec: 83.3% |

*Abbreviations: MAE = Mean Absolute Error; Acc = Accuracy; F1 = F1-score; Sens = Sensitivity; Spec = Specificity; AUC = Area Under the Curve.*

## 2.2. Data Imputation for Missing Records in EHR

The clinical utility of automated AHP diagnostic systems is limited by the incompleteness of patient records, where essential diagnostic variables are in some cases absent from structured EHR fields. Reliable assessment requires multiple interdependent variables, such as visual acuity, ocular motility findings, compensatory head positioning, and symptom reports that can be inconsistently documented across structured datasets. Moreover, many critical details, including diplopia characteristics, severity of postural adaptations, and functional impact, are frequently recorded in unstructured notes, making them inaccessible to conventional imputation methods focused on numerical data.

Data imputation in clinical domains has been extensively studied, with approaches ranging from traditional statistical models to advanced DL techniques. However, existing methods demonstrate limitations when applied to specialized diagnostic domains, such as AHP. Psychogyios et al. (2023) developed two DL models: the Improved Neighborhood Aware Autoencoder (I-NAA) and the Improved Generative Adversarial Imputation Network (I-GAIN), which outperformed traditional methods on four clinical datasets through batch normalization, variable K-Nearest Neighbor (k-NN) pre-imputation, and customized loss functions, achieving up to 9% higher F1-scores than baseline methods [10].

Building on GAN-based strategies, Bernardini et al. (2023) introduced a clinical conditional GAN (ccGAN) that incorporates demographic and clinical attributes into the imputation process. When evaluated on the multi-diabetic centers dataset and MIMIC-III, ccGAN improved imputation accuracy by 19.79% and predictive performance by up to 1.60%, demonstrating the value of embedding domain knowledge into imputation architectures [18]. Similarly, Weng et al. (2024) developed MVIIL-GAN, which combines autoencoder generators with variable- and instance-level discriminators to address both missingness and class imbalance, achieving a 5.4% AUC improvement over baselines and a 2.3% improvement over competing methods at 85% missingness on MIMIC-IV, even under extreme sparsity conditions [37].

Recent approaches have explored alternative imputation paradigms, but they face challenges related to interpretability. Liao et al. (2025) introduced Prompt as Pseudo-Imputation (PAI), which replaces missing entries with learnable prompts tailored for clinical prediction tasks. Evaluated on MIMIC-IV, CDSL, Sepsis, and eICU datasets, PAI outperformed both impute-then-regress and joint optimization baselines, achieving 4.1% AUPRC improvement on MIMIC-IV and up to 8.2% on Sepsis, with Transformer architectures showing superior performance [38]. Additionally, Firdaus et al. (2025) developed DRes-CNN, a residual convolutional architecture designed for high-missingness scenarios, which achieved an RMSE of 0.00006 on MIMIC-IV, representing an over 90% improvement compared to LL-CNN and U-Net [39].

Traditional ML approaches remain competitive but share the fundamental limitation of ignoring narrative clinical information. Ferri et al. (2023) demonstrated that simple imputers performed effectively in conjunction with classifiers such as random forest and gradient boosting to predict COVID-19 deaths, even when substantial data was missing [40]. Joel et al. (2024) reported that MissForest and MICE consistently yielded the lowest errors across multiple healthcare datasets [41]. In contrast, Chen et al. (2023) utilized Explainable Boosting Machines to highlight risks associated with imputation choices [42]. Similarly, Karimov et al. (2025) found that gradient boosting was the most reliable classifier for sarcopenia prediction across different imputation strategies. These findings suggest that, with appropriate imputation, classical ML pipelines remain competitive alternatives to more complex DL models [43].

Despite these advances, several gaps remain in EHR imputation research. Most existing approaches focus only on structured numerical variables and overlook the clinical notes available in patient records. Methods based on GANs and CNNs frequently demonstrate robust benchmark performance; however, they encounter difficulties in achieving consistent adaptability across datasets presenting different levels of missing information. Prompt-based methods, such as PAI, can improve prediction accuracy; however, their learned prompts lack a clear link to clinical variables, making them difficult to interpret. Traditional ML models remain competitive, but they rely on simple imputers and cannot capture the broader clinical context. To address these limitations, a curriculum learning–based imputation framework is proposed, which integrates both structured EHR variables and patient clinical notes. This approach integrates structured records with narrative information to produce imputations that capture both statistical patterns and clinical context, which can be more relevant to the clinical field. Table 2 presents a summary of recent studies on missing value imputation in EHR.

Table 2: A Summary of Recent Studies on Missing Value Imputation in EHR

| Study | Dataset | Model | Key Results |
|---|---|---|---|
| Psychogyios et al. (2023) | Framingham, Stroke, Physionet, UCI Heart | I-NAA, I-GAIN | F1-score improved by ~9% |
| Bernardini et al. (2023) | MDC, MIMIC-III | ccGAN | +19.8% imputation accuracy +1.6% predictive gain |
| Weng et al. (2024) | MIMIC-IV | MVIIL-GAN | AUC +5.4% vs baseline under 85% missingness |

| Liao et al. (2025) | MIMIC-IV, Sepsis, eICU | PAI (GRU/Transformer) | AUPRC +4.1% (MIMIC-IV) +8.2% (Sepsis) |
| --- | --- | --- | --- |
| Firdaus et al. (2025) | MIMIC-III/IV, Beijing Air Quality | DRes-CNN | ~92% error reduction |
| Ferri et al. (2023) | COVID-19 mortality (Madrid, Valencia) | k-NN, Bayesian ridge, GAN, RF, GBM | AUC 0.894 |
| Joel et al. (2024) | Breast cancer, heart disease, diabetes | Traditional ML + imputers | Accuracy up to 0.982 |
| Chen et al. (2023) | EHR benchmark datasets | EBMs + imputation | AUC range 69–99.9% |
| Pereira et al. (2023) | 34 medical datasets | PMIVAE | Superior MAE in 71% of datasets |
| Karimov et al. (2025) | Sarcopenia dataset | LR, RF, GBM, SVM with imputers | Accuracy up to 0.982 |

## 3. Proposed Methodology

### 3.1. Task 1- Diagnosis of Ocular-Induced Abnormal Head Posture Using Knowledge-based Image Analysis

This section outlines the proposed methodology for diagnosing ocular-induced AHP using a knowledge-based and image analysis framework. The integration of visual features from facial images with clinical attributes enables the framework to utilize multi-modal information for robust diagnosis. The processing pipeline consists of multiple components: data preparation, extraction of facial and ocular landmarks, head pose estimation, clinical feature extraction, and multi-level attention-based feature fusion using the proposed AHP-CADNet model. Each stage is designed to preserve essential anatomical cues and support the diagnosis of ocular misalignment and AHP assessment.

#### 3.1.1. Data Preparation

To support the learning process for diagnosis, the data preparation component is implemented. This includes multi-view facial image preprocessing, landmark extraction, head pose estimation, and extraction of clinically relevant features for each patient. All images are resized to a target width of 960 pixels, with the height adjusted to preserve the original aspect ratio, thereby maintaining the geometric relationships necessary for accurate assessment. The remaining data preparation steps are detailed in the following subsections.

##### 3.1.1.1. Face and Eye Region Landmarks Detection

The AHP-PoseGaze dataset [25] includes annotated images illustrating eye misalignment in the primary head position. These images are used to identify ocular landmarks for each patient. Eye landmark detection is performed using the MediaPipe Face Mesh framework [44], which provides 478 facial landmarks through a two-stage pipeline. This includes initial face localization using a face detector, followed by 3D regression to estimate fine-grained facial landmarks [44], [45]. For each eye, 16 dedicated landmarks and 5 iris landmarks are extracted, enabling precise localization of features relevant to eye misalignment detection, such as canthal positions, eyelid contours, and pupil centers. Based on detected landmarks, each eye region is segmented using an adaptive bounding box, with padding adjusted according to eye size to ensure full coverage of relevant features. The landmarks within each segmented region are then transformed from global face coordinates to local eye-region coordinates, spatially aligning the data for localized processing. This transformation reduces computational complexity while preserving the spatial relationships among key ocular features.

The iris region is identified based on five key landmarks: one central point and four perimeter points located at anatomically consistent positions, specifically the superior, inferior, nasal, and temporal boundaries. A least-squares circle fitting algorithm [46] is applied to the four perimeter points to estimate the iris boundary and refine the iris center, using:

$$\text{Residual}(x, y, r) = \sqrt{(x_i - x)^2 + (y_i - y)^2} - r \tag{1}$$

where $(x_i, y_i)$ are the perimeter points, $(x, y)$ is the estimated center, and $r$ is the fitted radius. This provided a consistent approximation of iris geometry across variable eye shapes. Subsequently, the pupil center is detected using intensity minima analysis within the segmented iris regions. The darkest pixel within the constrained iris boundary is identified using spatial averaging for noise reduction and sub-pixel refinement to achieve the precision necessary for detecting eye misalignment. Misalignment magnitude is then quantified as the Euclidean distance between the detected pupil center $(x_p, y_p)$ and the estimated iris center $(x_c, y_c)$:

$$\text{Misalignment Magnitude} = \sqrt{(x_p - x_c)^2 + (y_p - y_c)^2} \tag{2}$$

The directional deviation is calculated using:

$$\text{Angle} = \tan^{-1}\left(\frac{(y_p - y_c)}{(x_p - x_c)}\right) \times \left(\frac{180}{\pi}\right) \tag{3}$$

All extracted facial and ocular landmarks, including coordinates of the iris, eyelids, and canthi, are saved in structured JSON format to support the subsequent analysis. A visual summary of the eye landmark extraction process, including region segmentation, contour analysis, circle fitting, and final iris detection, is presented in Figure 1.

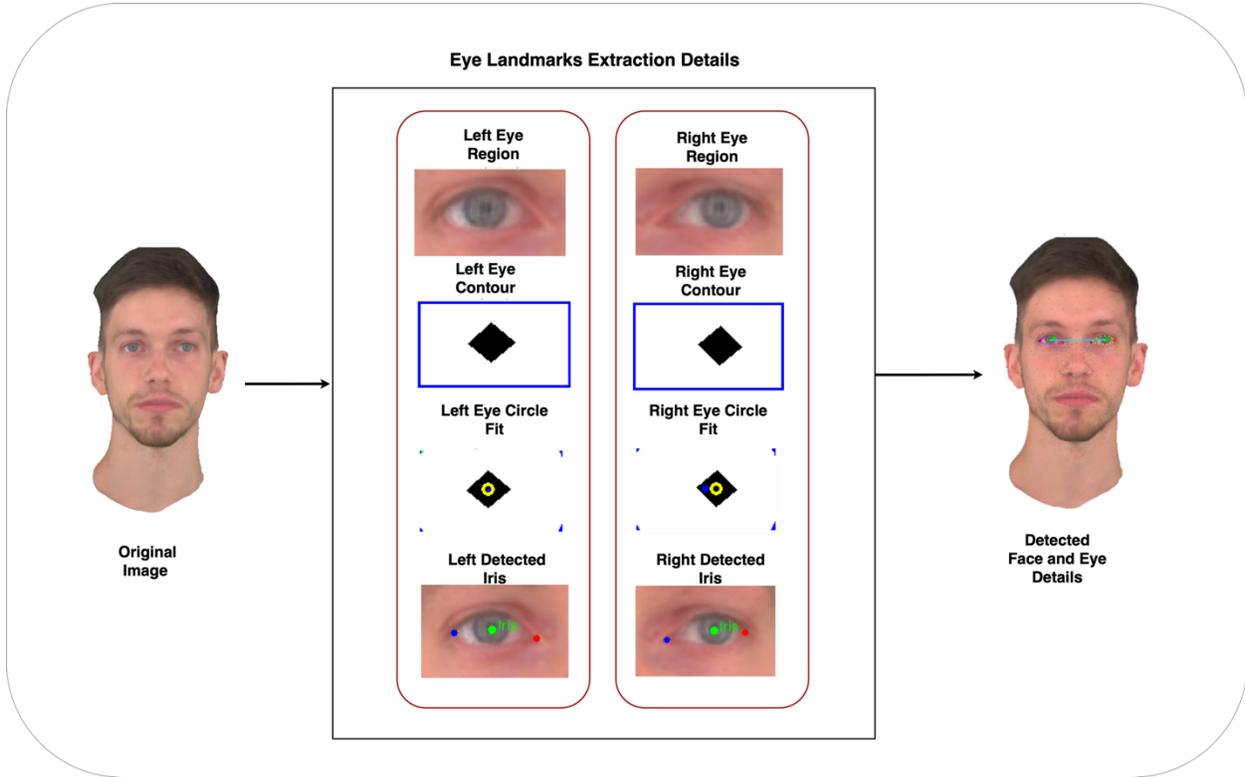

Figure 1: Visualization of facial and eye landmark detection

3.1.1.2.   Head Pose Landmarks Detection

Accurate evaluation of AHP requires a reliable estimation of head pose. Facial landmark-based methods have demonstrated robust performance in this context, due to their anatomical consistency and robustness to pose variation [47]. In this study, facial landmarks are extracted from seven pose-specific images per case in the PoseGaze-AHP dataset, corresponding to standardized views: frontal, up, down, left, middle-left, right, and middle-right. Building on the facial mesh generated in the previous stage using MediaPipe, a subset of anatomically informative landmarks, including those around the eyes, nose, mouth, chin, jawline, and facial contour, is used to infer head orientation.

The selected landmarks are represented in three coordinate formats: (i) normalized values scaled to the [0,1] range, (ii) absolute pixel positions within the image dimensions, and (iii) depth estimates (Z-axis) for spatial reasoning. A subset of 18 anatomically informative facial landmarks is selected for head pose estimation. These landmarks are further categorized into key anatomical groups: ocular (including the eye corners and center), nasal (encompassing the bridge, tip, and nostrils), oral (comprising the mouth corners and center), and craniofacial contour (including the chin, temples, jawline, and forehead). All extracted landmarks and metadata are stored in structured JSON format. An illustration of the seven standardized head pose views with detected landmarks is shown Figure 2.

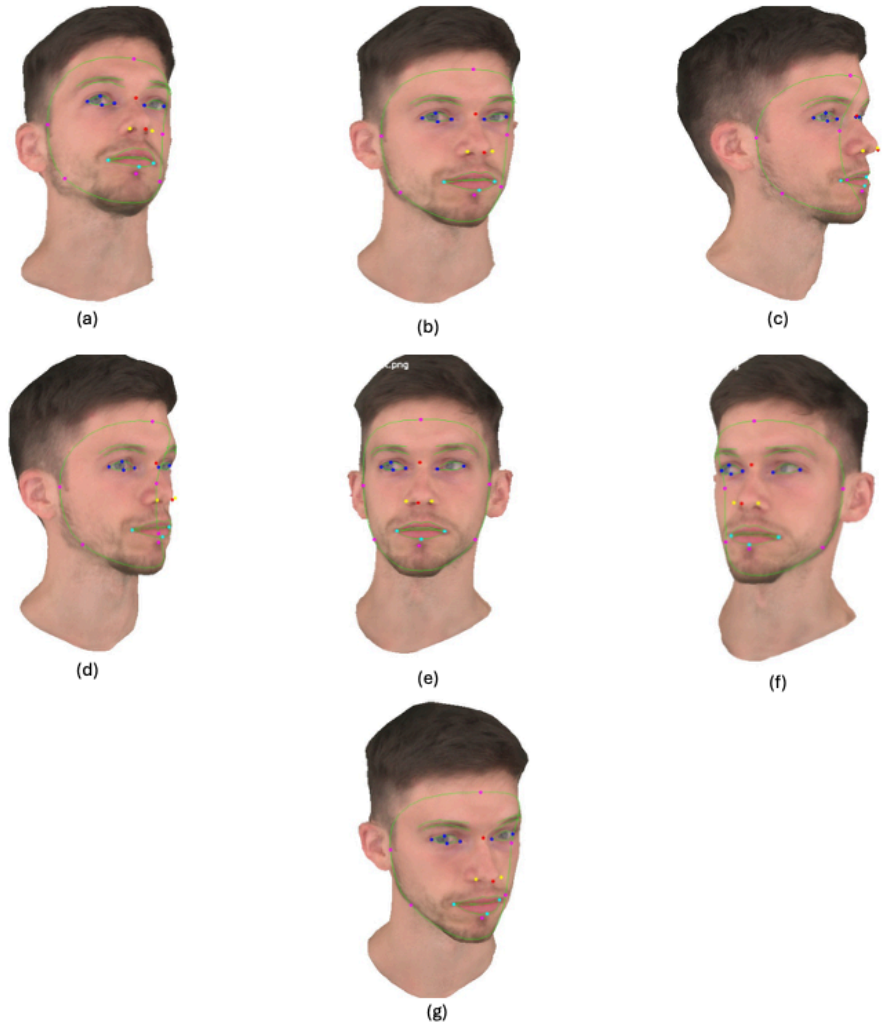

Figure 2: Detected facial landmarks across multi-views in PoseGaze-AHP dataset: (a) down, (b) frontal, (c) left, (d) middle-left, (e) middle-right, (f) right, and (g) up.

*3.1.1.3.    Descriptive Clinical Features Extraction*

The integration of structured and unstructured data as a multi-modal framework has been proven to enhance model performance. To incorporate this principle, descriptive clinical features were extracted from the PoseGaze-AHP dataset. This dataset was derived from a systematic review [26], in which all visualized cases are based on clinical information extracted from peer-reviewed research articles. Each indexed case is linked to its corresponding source publication. Clinical attributes are extracted for each case and integrated with the corresponding eye and head landmark data to enrich the diagnostic inputs. The extraction process started by identifying key clinical attributes from the dataset's metadata files, including diagnosis, AHP type, direction, and degree, diagnosed eye, associated eye misalignment type, and PD.

To generate complete clinical notes, the Claude Sonnet 4.0 language model [48] is prompted with both the structured patient features and the full publication content. Structured clinical notes are produced, from which relevant features are parsed. The resulting output preserves original clinical details while incorporating additional attributes such as patient demographics (age, gender), symptom duration, medical and surgical history, visual acuity, binocular vision status, ocular motility limitations, and head posture observations. A clinical feature extractor is then applied to convert these enriched notes into structured variables. Using pre-compiled regular expression patterns, the extractor identifies explicitly stated entities and computes derived values such as visual acuity asymmetry and motility deficit grades. In total, 32 discrete clinical features are extracted per case. These structured features are then merged with the corresponding eye and head pose landmarks to form a unified input representation for the proposed model architecture.

*3.1.2. Model Architecture Overview*

In this paper, the Abnormal Head Posture – Clinical Attention Diagnosis Network (AHP-CADNet) is proposed as a multi-level attention fusion framework for the diagnosis of ocular-induced AHP. The model integrates multiple data modalities, including eye and head landmark features as well as structured clinical attributes. Moreover, a multi-level attention mechanism is designed to capture both intra- and inter-modal relationships, thereby improving diagnostic performance. AHP-CADNet performs both classification and regression tasks. The classification tasks include identifying the AHP type, determining the affected eye, and diagnosing the underlying ocular condition. The regression tasks involve predicting prism diopter (PD) values and estimating the angular deviation of AHP. The architecture is organized into three multi-level attention fusion blocks, which progressively refine multimodal representations to enable reliable multi-output prediction. Furthermore, the framework is designed to reflect clinical progression by treating ocular misalignment as the primary task and AHP detection as a secondary task, consistent with the compensatory nature of AHP in response to underlying ocular conditions. The overall architecture of AHP-CADNet is illustrated in Figure 3.

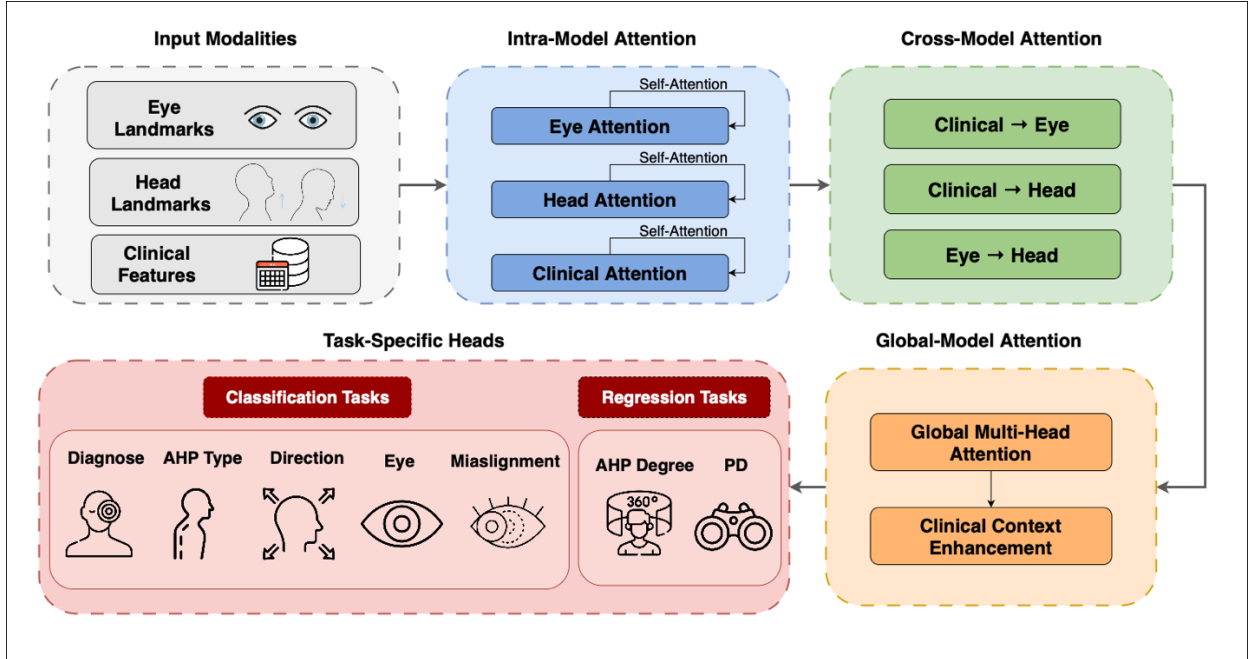

Figure 3: The Proposed AHP-CADNet Architecture

#### 3.1.2.1. Level 1: Intra-Modal Attention

In the first stage, each modality is independently preprocessed using a dedicated fully connected network. The input dimensions are reduced into modality-specific latent spaces, with 32 dimensions for eye features and 16 dimensions for head and clinical vectors. These reduced representations are then passed through intra-modal self-attention blocks, where each input is represented as a vector sequence, allowing self-attention to capture internal dependencies. Each block is implemented with multi-head self-attention using learned query, key, and value projections. In this manner, contextual relationships within each modality, such as interactions among eye landmarks or correlations between clinical indicators, are effectively modeled. The outputs are subsequently refined through residual connections and layer normalization to preserve gradient flow and stabilize the training process.

To enhance modality-specific interpretability, a feature importance gating mechanism is applied to each intra-modal output. Specifically, the gated representation is computed as:

$$z_{modality} = \text{Attention}(x) \odot \sigma(W^2 \cdot \text{ReLU}(W^1 \cdot x)) \qquad (4)$$

where $\sigma$ denotes the sigmoid activation function, and $W_1$, $W_2$ are learned weight matrices. This gating mechanism dynamically weights features according to their relevance, thereby improving both task performance and interpretability within each modality.

*3.1.2.2. Level 2: Cross-Modal Attention*

In addition to intra-modal modeling, multi-head attention is extended to cross-modal representation learning, following the Transformer architecture [49]. To enhance interpretability and control over inter-modal influence, a gated relevance mechanism is introduced to modulate cross-attention outputs based on a learned compatibility score between query and key features. Cross-modal interactions are modeled through attention modules that adopt the query–key–value formulation [50], where one modality serves as the query and another provides the keys and values. Specifically, clinical-to-eye attention is used to direct the model's focus toward diagnostically relevant eye representations, such as gaze patterns indicative of exotropia or hypertropia. Likewise, clinical-to-head attention is employed to emphasize particular head posture deviations (tilt, turn, chin-up, chin-down) based on corresponding clinical features. Moreover, an eye-to-head attention pathway is incorporated to capture anatomical correlations between ocular misalignment and AHPs, for instance, vertical gaze limitations resulting in a chin-down posture.

Each cross-modal module produces a context-aware representation that encodes dependencies between query and key–value features. Inspired by prior work on gated multimodal fusion [51] and cross-modal attention mechanisms [52], the framework incorporates a gated relevance mechanism in which a learned relevance score scales the attended representation. This score is derived by applying a feed-forward neural network to the concatenated query and key vectors, followed by a sigmoid activation. The gating mechanism enables adaptive control over the contribution of each cross-modal pathway according to its diagnostic importance. The relevance score $r$ and modulated attended output $z_{cross}$ are computed as follows:

$$r = \sigma(FFN([q; k])); \quad z_{cross} = CrossAttn(q, k, v) \cdot r \qquad (5)$$

where $\sigma$ denotes the sigmoid activation function, $[q; k]$ represents the concatenation of query and key vectors, and $FFN$ is a feed-forward network. This mechanism ensures that each cross-modal interaction is weighted according to its learned clinical relevance. Figure 4 illustrates the proposed gated relevance mechanism.

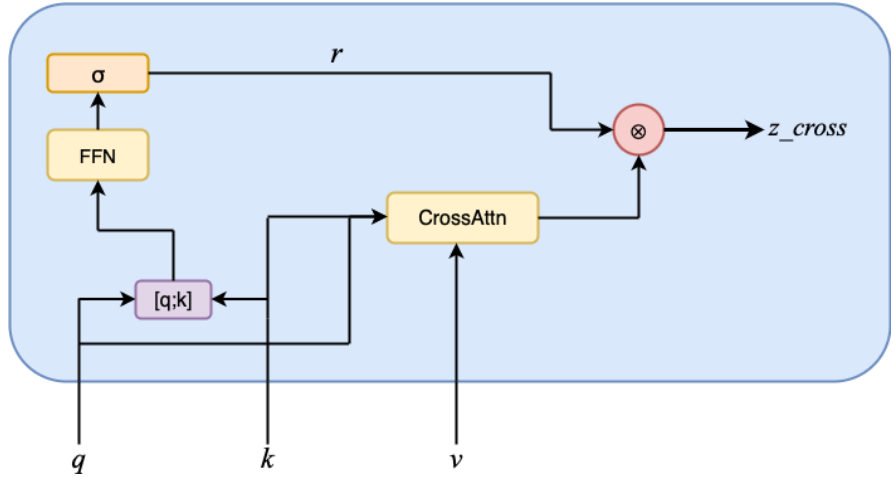

Figure 4: Gate Relevance Mechanism

### 3.1.2.3. Level 3: Global Context Attention

The third level performs global fusion across all attended streams, including the outputs of the clinical-to-eye, clinical-to-head, and eye-to-head attention modules, as well as the pure clinical features. Each of these streams is projected into a shared latent space, after which they are input into a multi-head attention block that models inter-modality dependencies. To explicitly regulate the contribution of each stream, a modality weighting network is employed. Attention-based weights are assigned through a learned projection followed by a softmax normalization:

$$\alpha = Softmax(W_{mod} \cdot [f^1; f^2; \dots ; f^M]) \quad (6)$$

where $\alpha \in \mathbb{R}^M$ denotes the modality weights, and $f_i$ are the projected representations of each modality stream. The modality-specific representations are then integrated through a weighted summation, producing a unified global representation. The resulting fused vector is passed through a clinical context enhancement network, which concatenates the global representation with the attended clinical vector. This combined representation is refined through a residual multi-layer perceptron (MLP) and normalized:

$$f_{enhanced} = LayerNorm(MLP([f_{global}; f_{clinical}]) + f_{global}) \quad (7)$$

The resulting enhanced feature vector is subsequently provided as input to a set of task-specific prediction heads. Each head is implemented as a lightweight MLP with batch normalization and dropout for regularization. The final architecture enables AHP-CADNet to model ocular misalignment as the primary task, encompassing diagnosis, affected eye, eye misalignment, and PD, while treating AHP detection, including type, direction, and degree, as a secondary task to reflect its compensatory role in clinical practice.

*3.2. Diagnosis and Clinical Imputation of Ocular-Induced AHP from Patient Notes*

In real-world clinical scenarios, incomplete clinical records are encountered due to limitations in documentation practices, variability in clinician reporting [53], omission of non-critical findings [54], and the prevalence of unstructured narrative notes. To address these challenges, a curriculum learning framework is proposed for imputing missing values in EHRs. The methodology is structured to include clinical description extraction, domain-specific data augmentation, and progressive learning schedules to ensure robustness, particularly for clinical conditions characterized by incomplete or partially observed data.

*3.2.1 Clinical Description Extraction*

Clinical information extraction follows the pipeline described in Section 3.1.1.3, utilizing the Claude Sonnet 4.0 language model, which is prompted with structured patient metadata and full-text clinical reports. Unlike the structured feature extraction described earlier, this stage focuses on generating concise textual summaries (150–200 words) for each indexed case. The generated notes capture the type, direction, and degree of AHP, the affected eye, and associated ocular misalignment, while also including contextual information such as demographics, symptom duration, history, visual acuity, binocular vision status, and treatment details. This process yields a total of 496 structured clinical notes, with an example illustrated in Figure 5.

> PDF: 155.pdf
> CaseID: case_0491
>
> Clinical Description:
>
> **A 4-year-old boy whose parents had noticed a right head tilt since the child was 1 year old presented to the clinic. His refractive error and best-corrected visual acuity (BCVA) were -9.25 DS/-4.50 DC at 180° and 20/60 in the right eye and -9.25 DS/-2.50 DC at 180° and 20/60 in the left eye. His near vision was N8 in both eyes. He had an esotropia of four prism diopters (PDs) for distance and near. Ocular motility evaluation showed an abduction deficit of -3 and grade 1 retraction with an up-shoot in the left eye on adduction. He also had INS with fine amplitude right beating nystagmus. A right head tilt of 30° was noted. Anterior segment examination was unremarkable in both the eyes. The retina was normal with myopic tessellations, and there was no torsion. An electroretinogram (ERG) revealed reduced amplitudes of both scotopic and photopic waveforms. The child was diagnosed with Duane retraction syndrome type 1 in the left eye, with INS and cone-rod dystrophy in both the eyes.**

Figure 5: A Sample of the Generated Clinical Notes

3.2.2. Domain-Specific Data Augmentation

Following the generation of clinical notes, a domain-specific augmentation approach is developed to overcome the constraints imposed by the limited number of clinical notes. Data augmentation in the medical field requires careful design to avoid introducing semantically incorrect or clinically invalid content [55]. General-purpose text augmentation methods often lack the constraints necessary for medical accuracy, which can lead to the alteration of critical diagnostic information. The proposed augmentation approach analyzes the previously generated clinical description to identify ophthalmological terminology, including diagnostic entities (e.g., superior oblique palsy, Duane syndrome), symptom notes, anatomical references, and measurement patterns. It also detects common reporting structures, directional terms (e.g., superior/inferior, medial/lateral), and standard abbreviations (e.g., OD for right eye), which are mapped to validated alternatives to ensure semantic consistency. The augmentation process employs four primary techniques:

- Domain-synonym substitution, where clinical terms are replaced with medically equivalent expressions (e.g., hypertropia → vertical deviation).
- Abbreviation variation, alternating between short forms and their expanded counterparts (e.g., OS → left eye).
- Clinical phrase rephrasing, modifying common reporting language without changing meaning (e.g., presented with → exhibited).
- Directional term variation, where anatomical directions are replaced with contextually appropriate alternatives when such substitutions do not alter clinical interpretation.

The augmentation approach is implemented utilizing Natural Language Toolkit (NLTK) [56] for text processing and regular expressions for pattern matching. The augmentation techniques are applied with a defined probability to ensure diverse and controlled variation across the dataset, as follows: domain-synonym substitution (50%), abbreviation variation (30%), phrase rephrasing (40%), and directional variation (15%). Each original case is expanded to four additional variants (n = 4), with contextual integrity preserved across all augmentations. A total of 1800 augmented notes are generated through this process, resulting in a combined dataset of 2296 clinical notes, including the original 496 cases. These notes are used as input to the model in the subsequent phase.

*3.2.1 Model Architecture Overview*

The proposed curriculum framework addresses missing clinical information by integrating progressive learning with domain-aware masking. It comprises four components: data preprocessing, a progressive masking strategy informed by domain knowledge, a Transformer-based model, and clinical target dependency modeling. Figure 6 illustrates the proposed curriculum learning framework.

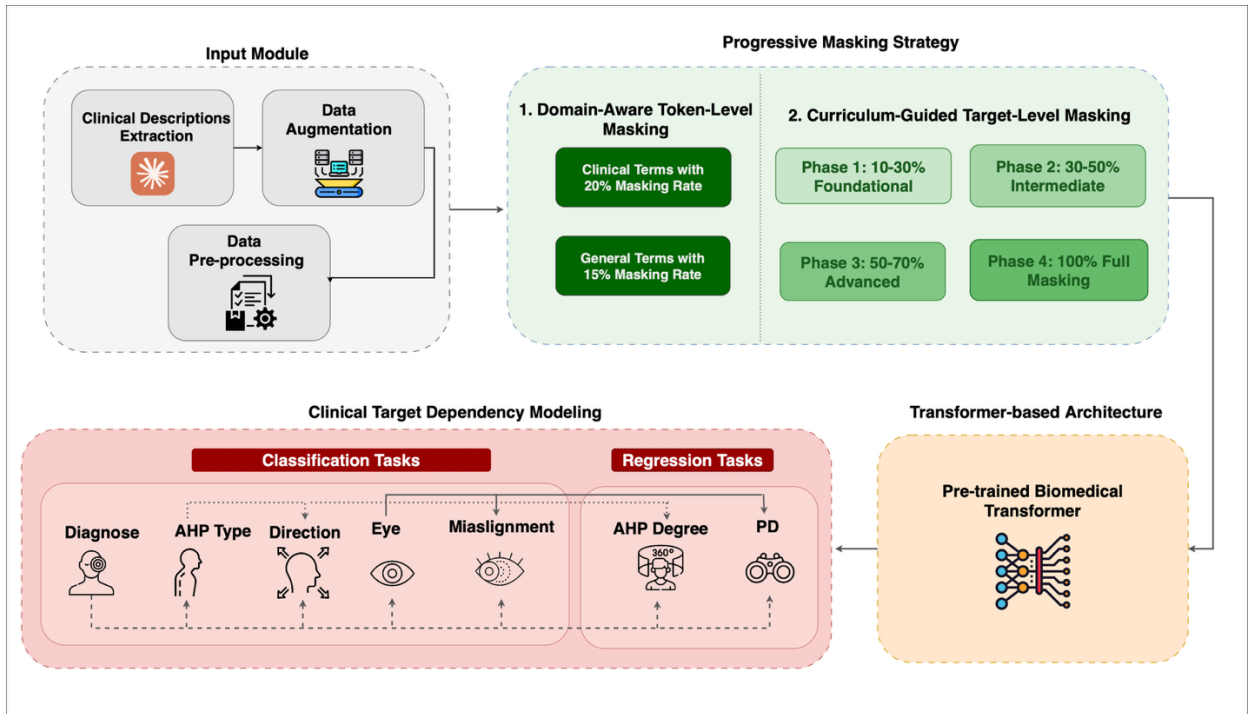

Figure 6: The Proposed Curriculum Learning Framework.

### 3.2.3.1. Data preprocessing

The dataset includes both original and augmented clinical notes, which are linked to corresponding targets within the dataset. Each target field is accompanied by an imputation indicator that specifies whether the value was originally missing. These indicators are essential for enabling controlled curriculum learning, as they allow selective masking of imputed targets during progressive training phases. To prepare the structured inputs for model consumption, appropriate encoding and normalization strategies are applied. Categorical variables are encoded using label encoding, while numerical variables are standardized to ensure consistent scaling across all prediction tasks.

### 3.2.3.2. Progressive Masking Strategy with Domain Awareness

The proposed architecture leverages a pretrained biomedical language model, which provides a foundation for understanding domain-specific terminology and contextual dependencies within clinical notes. These pretrained models serve as the backbone for both masked language modeling and multi-task diagnostic prediction. The proposed architecture introduces a comprehensive masking strategy at two key levels: (1) token-level masking, guided by domain-specific clinical knowledge and applied to extracted clinical notes, and (2) target-level masking, governed by a four-phase curriculum learning schedule that operates on the diagnostic targets and the records themselves. These mechanisms simulate real-world documentation uncertainty and support model generalization under varying degrees of data incompleteness.

A. *Domain-Aware Token-Level Masking:*

To enhance model robustness under conditions of clinical documentation variability, a domain-aware random masking strategy is applied to the extracted clinical notes. Clinically relevant terms, including diagnostic entities, anatomical references, and ocular misalignment descriptors, are identified using predefined domain-specific medical vocabularies. These tokens are assigned a higher masking probability of 25%, while general language tokens are masked at a lower rate of 10%. This differential masking mechanism encourages the model to focus on reconstructing diagnostic, meaningful information, thereby improving its capacity to infer missing values in critical clinical contexts.

B. *Curriculum-Guided Target-Level Masking:*

In parallel with token-level masking, a progressive masking strategy is applied to target fields, including both ocular misalignment and AHP labels. This approach incrementally increases task difficulty across four curriculum phases, allowing the model to gradually adapt to more challenging imputation scenarios. At each phase, initially imputed target values are selectively masked based on a predefined schedule, guiding the model through a structured learning progression. Table 3 illustrates the progressive phases of curriculum learning, along with the associated masking rates.

Table 3: Curriculum Learning Phases

| Phase | Masking Rate | Description |
|---|---|---|
| Phase 1 – Foundational Learning | 10% – 30% | Model is trained on mostly complete inputs to learn clinical patterns and terminology under minimal sparsity. |
| Phase 2 – Intermediate Training | 30% – 60% | Encourages the model to perform multi-target imputation using partially observed contextual cues. |
| Phase 3 – Advanced Training | 60% – 90% | Exposes the model to highly incomplete inputs, promoting generalization and pattern inference under sparse conditions. |
| Phase 4 – Full Masking Training | 100% | Requires the model to infer all diagnostic targets without any observed labels, simulating the deployment setting. |

This curriculum masking strategy supports a structured transition from complete to sparse clinical inputs. Figure 7 illustrates a sample of token- and target-level masking strategy.

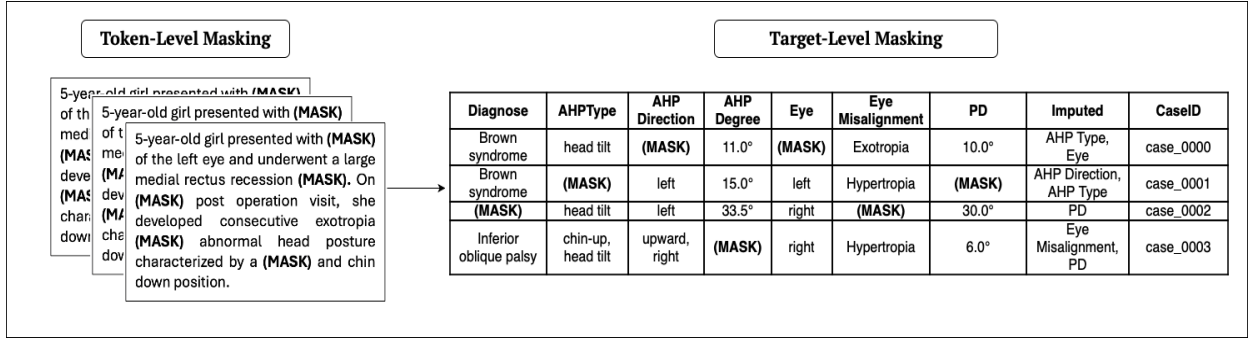

*Figure 7: Illustration of token- and target-level masking strategy.*

### 3.2.3.3 Transformer-Based Imputation Architecture

Building on domain-aware masking and progressive curriculum strategies, a transformer-based neural architecture is implemented for clinical imputation and diagnostic prediction. The architecture is designed to learn contextual language representations and structured diagnostic targets from incomplete clinical notes. It comprises four main components: a transformer-based language encoder, a shared clinical feature extraction module, task-specific prediction heads, and a clinical target dependency modeling module.

The core of the architecture employs a pretrained biomedical transformer encoder to process descriptive input. Several domain-specific language models are evaluated, including PubMedBERT [57], BioBERT [57], and SciBERT [58], to identify the most suitable backbone for ophthalmology-related language representation. The selected encoder tokenizes and encodes clinical notes, generating embeddings that are forwarded to two parallel branches: a masked language modeling (MLM) head for token reconstruction and a diagnostic prediction module. The diagnostic branch incorporates a shared feature extraction module composed of two fully connected layers with ReLU activation and dropout. These modules feed seven task-specific prediction heads: five for classification tasks (diagnose, AHP type, AHP direction, affected eye, and eye misalignment) and two for regression tasks (AHP degree and PD). The training process jointly optimizes the MLM and diagnostic objectives through a weighted multi-objective loss function:

$$L_{total} = 0.2 \times L_{MLM} + 0.8 \times L_{diagnostic} \tag{8}$$

The diagnostic loss component is computed based on the curriculum schedule. Loss is only applied to targets that are either originally complete or not masked during the current curriculum phase, with dependency-based weighting to reflect clinical relationships:

$$L_{diagnostic} = \left(\frac{1}{N}\right) \times \sum_{i=1}^{N} L_{i(\hat{y}_i, y_i)} \times w_{dep}^i \times \mathbb{1}[(\neg imputed_i) \vee (\neg masked_{curriculum_i})] \quad (9)$$

where $L_i$ denotes cross entropy for classification targets and mean squared error (MSE) for regression tasks, and $w_{dep}^i$ represents dependency-based weighting reflecting clinical relationship strength for target $i$. A curriculum-aware early stopping mechanism is enforced, ensuring a minimum of 25 epochs at 100% masking to guarantee full imputation readiness before deployment.

*3.2.3.4 Clinical Target Dependency Modeling*

To capture domain-specific clinical relationships between imputation targets, interdependencies among the targets are modeled based on diagnostic hierarchies. The design assumes that certain clinical features serve as predictors for others. In this context, the diagnosis label functions as a complete predictor (never imputed) and informs all other predictions. The dependency relationships are applied using attention-based mechanisms. Each target $i$ first generates a target-specific embedding:

$$f_i^{\{target\}} = TargetEmbedding_i(f_{\{shared\}}) \quad (10)$$

Clinical dependencies are captured through learned attention weights:

$$\alpha_{\{j \to i\}} = \sigma\left(W_{\{dep\}} \cdot \left[f_j^{\{target\}}; f_i^{\{target\}}\right]\right) \quad (11)$$

The final enhanced representation for each target incorporates information from its dependent source targets:

$$f_i^{\{enhanced\}} = f_i^{\{target\}} + \sum_{\{j \in D(i)\} \alpha_{\{j \to i\}}} \alpha_{\{j \to i\}} \cdot f_j^{\{target\}} \quad (12)$$

where $f_{\{shared\}} \in \mathbb{R}^{256}$ is the shared clinical feature representation, $f_i^{\{target\}}$ is the target-specific embedding for clinical target $i$, $\alpha_{\{j \to i\}}$ represents the attention weight from source target $j$ to dependent target $i$, $W_{\{dep\}}$ is the learned dependency attention matrix, $\sigma$ represents the sigmoid activation function, and $D(i)$ denotes the set of source targets that influence target $i$ according to the clinical dependency structure defined in Table 4.

Table 4: Clinical Target Dependency Structure

| Source Target | Dependent Targets |
| --- | --- |
| Diagnose | AHP Type, AHP Direction, Eye, Eye Misalignment, AHP Degree, PD |
| AHPType | AHP Direction, AHP Degree |
| AHPDirection | AHP Degree |
| Eye | Eye Misalignment, PD |
| EyeMisalignment | PD |

Clinical target dependencies are modeled using a cross-target attention mechanism, enabling contextual interaction among related labels. Each target incorporates information from its clinically relevant dependencies via learned attention weights, allowing the model to prioritize more informative features. Diagnose, as a fully observed label, is weighted more heavily and guides the overall prediction structure. Shared clinical representations are transformed into target-specific embeddings using lightweight projection layers, followed by cross-target attention to aggregate dependency-aware features. This design preserves clinically meaningful reasoning patterns during the imputation process.

## 4 Dataset

The PoseGaze-AHP dataset [25] is a 3D knowledge-based collection of simulated cases, organized into two primary parts for each case. The first part focuses on ocular misalignment and includes a single image captured in the primary head position, highlighting the misaligned eye. This image is associated with a metadata file that consists of the diagnosis type, classification of eye misalignment, the affected eye, and the deviation measured in PD. The second part focuses on AHP and comprises seven images representing compensatory head positions, including different viewpoints: Frontal, Left, Right, Up, Down, Middle_Left, and Middle_Right. The corresponding metadata provides information on the AHP type, direction, and the angular degree of deviation. The dataset comprises 496 clinical cases, each rendered with two primary mask textures, resulting in a total of 9920 images. Data was originally imputed using medical imputation rules derived from the studied research papers in a previously published systematic review [26].

## 5 Experimental Settings

All experiments are conducted in a Google Colab environment equipped with an NVIDIA T4 GPU. For AHP-CADNet, the dataset is divided into 4,788 images for training (70%), 1,026 for validation (15%), and 1,027 for testing (15%). The fusion configuration uses intra-modal dimensions of [32, 16, 16], a cross-modal dimension of 32, a global dimension of 64, four attention heads, and a dropout rate of 0.1. Training runs for 150 epochs with early stopping patience of 15. Optimization is performed with AdamW, and a ReduceLROnPlateau scheduler adjusts the learning rate with a factor of 0.7.

For the curriculum learning–based imputation framework, a total of 2,296 clinical notes are used, with 1,377 allocated for training (60%), 459 for validation (20%), and 459 for testing (20%). The framework is implemented with transformer backbones and trained for 150 epochs under a curriculum scheduler that progressively increases the masking rate from 0.1 to 1.0, with a guaranteed capability phase enforced during the final 25 epochs. Training uses AdamW optimization, together with curriculum-aware early stopping patience of 15.

## 6 Evaluation Measures

The performance of the proposed frameworks is evaluated using a combination of classification and regression metrics to ensure a comprehensive assessment of diagnostic reliability. For classification tasks, accuracy is used to quantify the overall proportion of correctly predicted samples. The F1-score, on the other hand, provides a balanced measure of precision and recall, particularly useful in handling class imbalance. Sensitivity reflects the model's ability to correctly identify positive cases, whereas specificity measures its ability to recognize negative cases, thereby reducing false positives.

For regression tasks, performance is assessed using the MAE, which captures the average magnitude of prediction errors, and the coefficient of determination ($R^2$), which indicates the proportion of variance in the ground truth explained by the model. These measures enable robust validation of diagnostic accuracy and clinical applicability.

## 7 Experimental Results

*6.1 AHP-CADNet Framework Evaluation*

To assess the diagnostic performance of AHP-CADNet, a set of comparative experiments is conducted across multiple tasks, including ocular classification, misalignment detection, PD regression, ocular diagnosis classification, and AHP characterization (type, direction, and degree). The results, summarized in Table 5, report classification and regression performance across a range of model variants.

Table 5: Comparative Experimental Results

| Model | Task | Accuracy | F1 | Sensitivity | Specificity | MAE | $R^2$ | Correlation |
|---|---|---|---|---|---|---|---|---|
| | | | **Image Key points** | | | | | |
| ViT | Eye | 98.6% | 98.4% | 98.1% | 99.1% | | | |

| Model | Task | | | | | | | |
|---|---|---|---|---|---|---|---|---|
| | Eye Misalignment | 90.8% | 56.2% | 54.3% | 99% | | | |
| | PD | | | | | 0.175 | 0.923 | 0.963 |
| | Diagnose | 72.8% | 34.4% | 34% | 98.9% | | | |
| | AHP Type | 96.8% | 79% | 76% | 99.6% | | | |
| | AHP Direction | 96.5% | 74.1% | 72.2% | 99.7% | | | |
| | AHP Degree | | | | | 0.112 | 0.939 | 0.97 |
| **Image Key points & Clinical Information** | | | | | | | | |
| ViT | Eye | 98.2% | 97.9% | 97.2% | 98.8% | | | |
| | Eye Misalignment | 91% | 54.1% | 53% | 99.1% | | | |
| | PD | | | | | 0.174 | 0.92 | 0.961 |
| | Diagnose | 70.5% | 32.7% | 32.3% | 98.8% | | | |
| | AHP Type | 95.5% | 72.5% | 70.6% | 95.5% | | | |
| | AHP Direction | 95.6% | 59.7% | 56.9% | 95.6% | | | |
| | AHP Degree | | | | | 0.113 | 0.936 | 0.968 |
| AHP-CADNet | Eye | 98.5% | 98.2% | 97.9% | 99.2% | | | |
| | Eye Misalignment | 97.5% | 85.5% | 83.2% | 99.7% | | | |
| | PD | | | | | 0.199 | 0.933 | 0.969 |
| | Diagnose | 96.9% | 92.3% | 91.3% | 99.9% | | | |
| | AHP Type | 98.3% | 97.1% | 96.1% | 99.8% | | | |
| | AHP Direction | 99% | 95.3% | 92.8% | 99.9% | | | |
| | AHP Degree | | | | | 0.103 | 0.922 | 0.961 |
| AHP-CADNet (Early Fusion) | Eye | 94.4% | 93% | 94.4% | 0.944 | | | |
| | Eye Misalignment | 92.2% | 67.6% | 92.2% | 0.922 | | | |
| | PD | | | | | 0.635 | 0.475 | 0.717 |
| | Diagnose | 90.8% | 56.8% | 90.8% | 90.8% | | | |
| | AHP Type | 87.8% | 44.9% | 87.8% | 87.8% | | | |
| | AHP Direction | 88% | 35% | 88% | 88% | | | |
| | AHP Degree | | | | | 0.432 | 0.297 | 0.574 |
| AHP-CADNet (Late Fusion) | Eye | 90% | 89.1% | 90% | 90% | | | |
| | Eye Misalignment | 87.7% | 55.4% | 87.7% | 87.7% | | | |
| | PD | | | | | 0.636 | 0.431 | 0.716 |
| | Diagnose | 7.83% | 32% | 7.83% | 7.83% | | | |
| | AHP Type | 88.4% | 44.6% | 88.4% | 88.4% | | | |
| | AHP Direction | 89.1% | 32.5% | 89.1% | 89.1% | | | |
| | AHP Degree | | | | | 0.430 | 0.298 | 0.596 |
| AHP-CADNet (without | Eye | 85.2% | 82.4% | 80.4% | 91.3% | | | |
| | Eye Misalignment | 88.4% | 52.4% | 49.5% | 98.7% | | | |

| Model | Task | Acc | F1 | Sens | Spec | MAE | PCC | CCC |
|---|---|---|---|---|---|---|---|---|
| Intra-Modal Attention) | PD | | | | | 0.385 | 0.713 | 0.881 |
| | Diagnose | 79.4% | 34.3% | 35.5% | 99.2% | | | |
| | AHP Type | 93.9% | 77.4% | 72.7% | 99.2% | | | |
| | AHP Direction | 94.4% | 57.3% | 55.3% | 99.4% | | | |
| | AHP Degree | | | | | 0.178 | 0.872 | 0.958 |
| AHP-CADNet (Eye Landmarks & Clinical Features) | Eye | 97.2% | 96% | 94.9% | 98.4% | | | |
| | Eye Misalignment | 98.1% | 87.1% | 86.2% | 99.8% | | | |
| | PD | | | | | 0.217 | 0.922 | 0.963 |
| | Diagnose | 91.8% | 61.6% | 60.1% | 99.7% | | | |
| | AHP Type | 94.8% | 82.2% | 81% | 99.3% | | | |
| | AHP Direction | 92% | 62.4% | 60.6% | 99.1% | | | |
| | AHP Degree | | | | | 0.228 | 0.76 | 0.874 |
| AHP-CADNet (Head Landmarks & Clinical Features) | Eye | 84% | 82% | 80.5% | 90.6% | | | |
| | Eye Misalignment | 87.2% | 63.9% | 57.7% | 98.6% | | | |
| | PD | | | | | 0.479 | 0.658 | 0.814 |
| | Diagnose | 87.4% | 53.1% | 51.9% | 99.5% | | | |
| | AHP Type | 98.1% | 93.1% | 90.1% | 99.7% | | | |
| | AHP Direction | 97.1% | 75.2% | 70.6% | 99.7% | | | |
| | AHP Degree | | | | | 0.123 | 0.934 | 0.967 |
| AHP-CADNet (Clinical Features) | Eye | 64.8% | 50.8% | 50.9% | 77.7% | | | |
| | Eye Misalignment | 72.9% | 25.5% | 26.4% | 96.9% | | | |
| | PD | | | | | 0.514 | 0.601 | 0.797 |
| | Diagnose | 65.4% | 20.2% | 21.3% | 98.6% | | | |
| | AHP Type | 77.9% | 22.3% | 23% | 96.7% | | | |
| | AHP Direction | 67.5% | 10.5% | 11% | 95.4% | | | |
| | AHP Degree | | | | | 0.273 | 0.7 | 0.844 |

The experimental results demonstrate that the complete AHP-CADNet consistently outperforms all model variants across both classification and regression tasks. High performance is achieved in eye misalignment with 97.5% accuracy and 85.5% F1-score, in diagnosis with 96.9% accuracy and 92.3% F1-score, and in AHP detection with 98.3% accuracy for type prediction and 99.0% for direction, while maintaining robust sensitivity and specificity. For regression, the model exhibits low MAE of 0.199 for prism diopters and 0.103 for AHP degree, accompanied by strong correlation coefficients of 0.969 and 0.961, respectively. These findings highlight the effectiveness of hierarchical multimodal fusion and attention mechanisms in delivering reliable diagnostic predictions. The baseline Vision Transformer (ViT), which includes eye and head pose embeddings with a clinical token, achieves competitive performance with 98.6% accuracy in eye classification. However, it shows limited capacity to integrate multimodal inputs effectively. Although the inclusion of clinical features yields a slight improvement in eye misalignment detection, increasing accuracy from 90.8% to 91.0%, performance in diagnosis decreases from 72.8% to 70.5%, and results in AHP-related tasks are also reduced. This reduction is likely due to the absence of modality-specific encoding and attention-based fusion in the ViT architecture.

The early fusion variant of AHP-CADNet, which concatenates all input modalities before processing, achieves moderate performance on simple tasks but degrades on more complex ones. For instance, diagnosis classification reaches 90.8% accuracy but only 56.8% F1-score, while PD regression yields a high error with an MAE of 0.635. The late fusion variant, which encodes each modality independently before merging, performs even worse across most metrics, indicating that neither strategy adequately captures inter-modal dependencies. In contrast, the complete AHP-CADNet with multi-level attention modules consistently outperforms all other variants, achieving 97.5% accuracy and 85.5% F1-score in eye misalignment, 96.9% accuracy and 92.3% F1-score in diagnosis, and strong results across all AHP-related tasks. Its regression performance is also robust, with MAEs of 0.199 for PD and 0.103 for the AHP degree, validating the effectiveness of multi-level attention-based fusion.

Another experiment in which intra-modal attention is removed from AHP-CADNet and replaced with lightweight feedforward modules results in substantial performance degradation. Eye classification accuracy decreases from 98.5% to 85.2%, and diagnosis F1-score drops from 92.3% to 34.3%, while regression errors for both PD and AHP degree notably increased. These results confirm the importance of modeling fine-grained intra-modal patterns prior to cross-modal integration. To further assess the contribution of each modality, three restricted AHP-CADNet variants are evaluated. The eye-and-clinical model, which excludes head pose features, performs well on ocular tasks, achieving an F1-score of 0.871 for eye misalignment and 0.616 for diagnosis; however, it underperforms on AHP-related regression, with an AHP degree MAE of 0.228. Conversely, the head-and-clinical model, which excludes eye landmarks, performs better in AHP-related tasks but performs less effectively on eye-specific predictions. The clinical-only model, which excludes all anatomical data, yields the lowest results overall, with a diagnosis F1-score of 0.202 and a high PD MAE of 0.514, confirming that structured clinical data alone is insufficient for reliable prediction.

To summarize, the experimental results validate the AHP-CADNet architecture, showing that its multi-level fusion strategy delivers robust diagnostic performance by leveraging the complementary strengths of eye, head, and clinical features. In contrast, simplified fusion strategies and reduced attention mechanisms consistently lead to inferior outcomes, while modality-restricted variants highlight the distinct contributions of each input source.

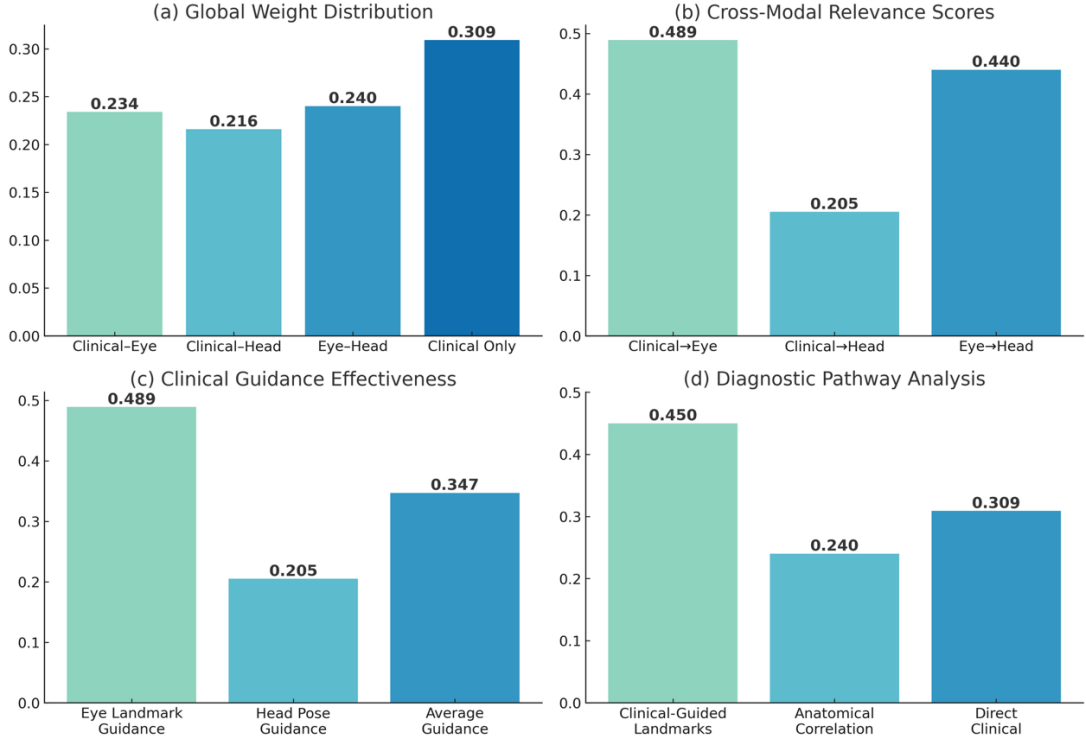

Figure 8: Multimodal Interpretability Analysis of AHP-CADNet

Figure 8 presents a multimodal interpretability analysis of AHP-CADNet, illustrating how clinical, ocular, and postural features contribute to the diagnostic reasoning process. In Fig. 8a, the relative importance of each fusion type is quantified based on global attention weights. Although clinical information receives the highest overall attention weight (0.309), its standalone predictive performance remains the lowest among all variants, as reported in Table 1, where merging eye and head landmarks captures compensatory patterns critical for AHP characterization. Among fused modality pairs, Eye–Head (0.240) is assigned the highest weight, followed by Clinical–Eye (0.234) and Clinical–Head (0.216), indicating that anatomical cues dominate when clinical input is less directly informative. Figure 8b evaluates cross-modal interaction strength through relevance scores computed between modality pairs. Clinical-to-Eye pathways exhibit the highest influence (0.489), followed by Eye-to-Head (0.440), while Clinical-to-Head interactions are substantially weaker (0.205). This asymmetry indicates that clinical descriptors align more closely with ocular abnormalities, which is consistent with the dataset, as the clinical features are extracted from textual descriptions in ophthalmology literature that primarily emphasize ocular aspects rather than head pose.

Figure 8c evaluates the effectiveness of clinical guidance, measuring how well clinical features enhance the learning of eye and head representations. The model shows stronger guidance to eye features (0.489) compared to head pose features (0.205), reinforcing the interpretation that clinical inputs more directly inform ocular predictions. Figure 8d aims to decompose the diagnostic reasoning into distinct functional pathways. The most influential pathway is Clinical-Guided Landmark interpretation (45.0%), followed by Direct Clinical inference (30.9%) and Anatomical Correlation (24.0%). These findings suggest that AHP-CADNet benefits most when clinical features guide anatomical interpretation, rather than acting as isolated predictors. Together, the four subfigures provide compelling evidence that AHP-CADNet not only integrates multimodal information effectively but also leverages structured attention to prioritize clinically meaningful and interpretable decision routes.

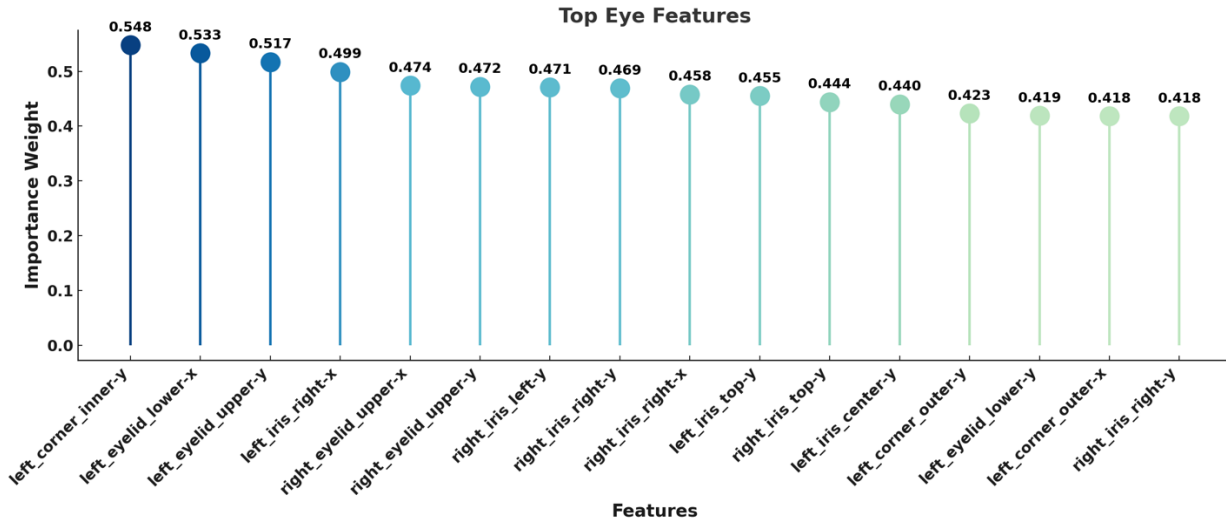

(a)

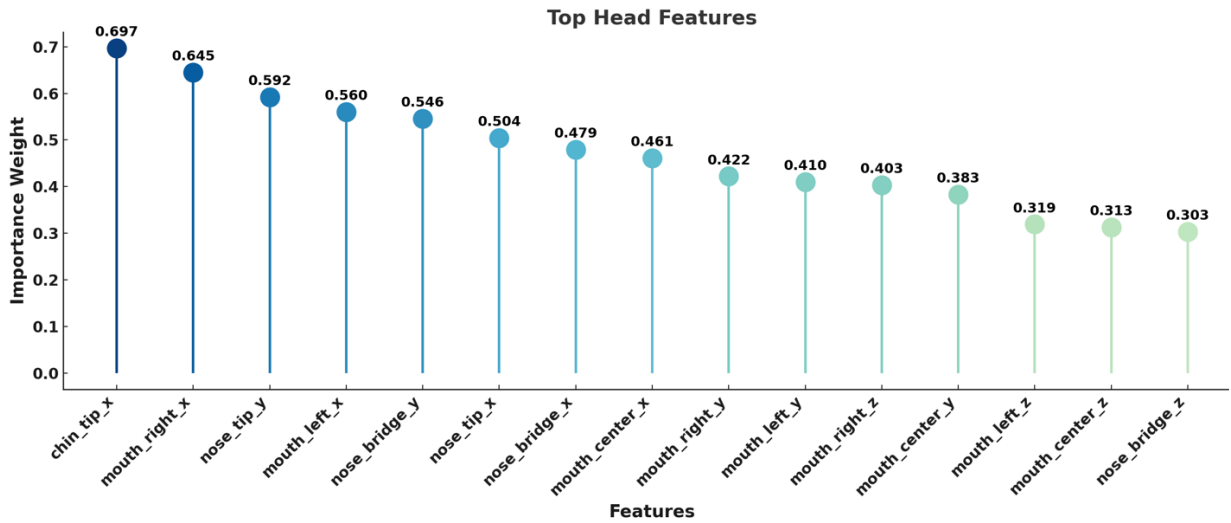

(b)

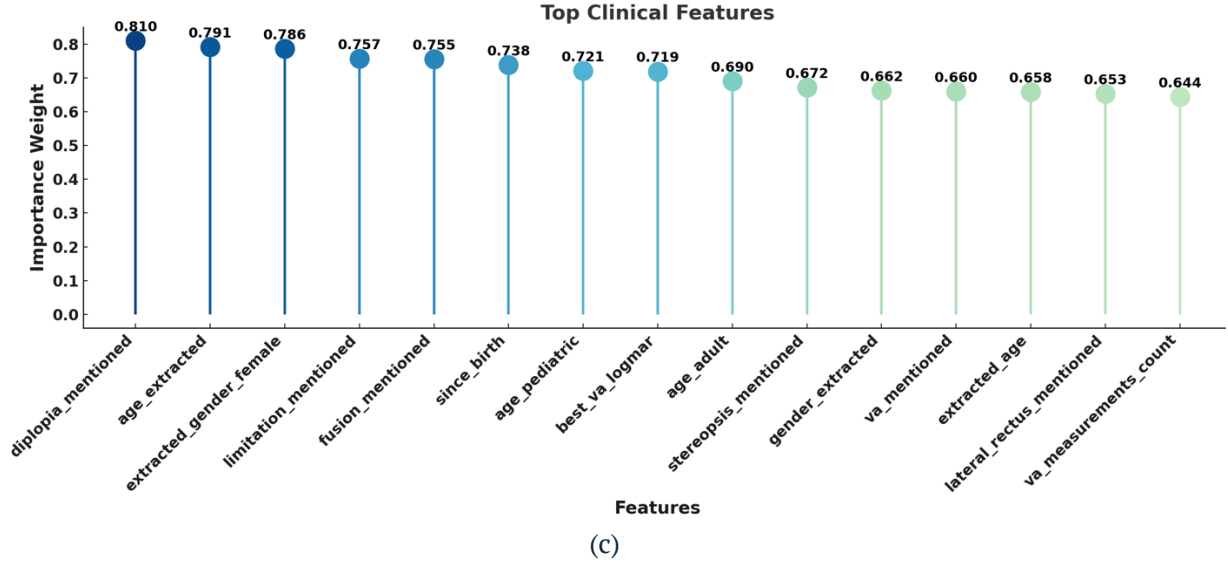

(c)

Figure 9: Feature Importance Analysis Across Modalities

Figure 9 presents an analysis of the top-ranked features within each modality, as determined by attention-based importance weights. For ocular features (Fig. 9a), the most influential variables include left corner inner y (0.548), right eyelid lower y (0.533), and left eyelid lower x (0.517). These features capture the geometric configuration of eyelid and iris landmarks, underscoring their role in modeling ocular misalignment. For AHP estimation (Fig. 9b), the most informative variables are chin tip x (0.697), mouth right x (0.645), and nose tip y (0.592). These highlight the diagnostic relevance of lower facial and midline craniofacial structures in detecting AHP. The inclusion of both horizontal and vertical displacement measures further reflects the multidimensional nature of head orientation analysis. In the clinical domain (Fig. 9c), the top-ranked features are diplopia mentioned (0.810), age extracted (0.791), and gender extracted (0.786). Additional influential variables include the number of VA measurements (count of visual acuity records), the best VA logMAR (best recorded visual acuity in logMAR units), and whether fusion is mentioned (whether binocular fusion is reported), demonstrating the model's ability to prioritize structured symptom descriptors alongside patient metadata.

## 6.2 Curriculum Learning Framework Evaluation

The experimental evaluation in this section is divided into two parts: (1) curriculum learning using descriptive data to predict all targets, including both complete and imputed values; and (2) curriculum learning applied specifically to the prediction of imputed values. Table 6 presents a summary of all results obtained.

Table 6: Experimental Results of Curriculum Learning Framework.

| Model | Task | Accuracy | F1 | Sensitivity | Specificity | MAE | R² | Correlation |
|---|---|---|---|---|---|---|---|---|
| | | **Overall Prediction** | | | | | | |
| BioBERT | Diagnose | 97.17% | 97.02% | 97.17% | 96.87% | | | |
| | AHPType | 99.13% | 99.21% | 99.13% | 99.14% | | | |
| | AHPDirection | 96.08% | 95.32% | 96.08% | 95.41% | | | |
| | Eye | 92.16% | 92.55% | 92.16% | 92.10% | | | |
| | EyeMisalignment | 98.69% | 98.91% | 98.69% | 98.70% | | | |
| | AHPDegree | | | | | 0.277 | 0.622 | 0.7696 |
| | PD | | | | | 0.266 | 0.703 | 0.8264 |
| SciBERT | Diagnose | 97.60% | 96.95% | 97.60% | 97.14% | | | |
| | AHPType | 99.78% | 99.80% | 99.78% | 99.78% | | | |
| | AHPDirection | 96.30% | 95.00% | 96.30% | 95.51% | | | |
| | Eye | 93.46% | 93.26% | 93.46% | 93.27% | | | |
| | EyeMisalignment | 97.17% | 97.54% | 97.17% | 97.22% | | | |
| | AHPDegree | | | | | 0.283 | 0.576 | 0.7732 |
| | PD | | | | | 0.280 | 0.690 | 0.7992 |
| PubMed BERT | Diagnose | 97.17% | 96.90% | 97.17% | 96.67% | | | |
| | AHPType | 99.78% | 99.80% | 99.78% | 99.78% | | | |
| | AHPDirection | 96.51% | 96.04% | 96.51% | 96.02% | | | |
| | Eye | 93.46% | 93.33% | 93.46% | 93.31% | | | |
| | EyeMisalignment | 97.17% | 97.71% | 97.17% | 97.14% | | | |
| | AHPDegree | | | | | 0.249 | 0.639 | 0.7724 |
| | PD | | | | | 0.267 | 0.734 | 0.8101 |
| | | **Imputed Only Prediction** | | | | | | |
| BioBERT | Diagnose | - | - | - | - | - | - | |
| | AHPType | 100.00% | 100.00% | 100.00% | 100.00% | | | |
| | AHPDirection | 92.06% | 93.08% | 92.06% | 91.68% | | | |
| | Eye | 85.34% | 91.36% | 85.34% | 87.44% | | | |
| | EyeMisalignment | 87.50% | 93.75% | 87.50% | 88.69% | | | |
| | AHPDegree | | | | | 0.125 | 0.820 | 0.8658 |
| | PD | | | | | 0.056 | 0.941 | 0.8006 |
| SciBERT | Diagnose | - | - | - | - | - | - | |
| | AHPType | 100.00% | 100.00% | 100.00% | 100.00% | | | |
| | AHPDirection | 92.06% | 89.45% | 92.06% | 90.36% | | | |
| | Eye | 88.36% | 89.45% | 88.36% | 88.65% | | | |
| | EyeMisalignment | 75.00% | 91.67% | 75.00% | 77.08% | | | |
| | AHPDegree | | | | | 0.123 | 0.842 | 0.9180 |
| | PD | | | | | 0.059 | 0.950 | 0.8514 |
| PubMed BERT | Diagnose | - | - | - | - | - | - | |
| | AHPType | 100.00% | 100.00% | 100.00% | 100.00% | | | |
| | AHPDirection | 94.71% | 94.95% | 94.71% | 94.20% | | | |
| | Eye | 87.93% | 90.42% | 87.93% | 88.74% | | | |

| | | | | | | | |
|---|---|---|---|---|---|---|---|
| EyeMisalignment | 75.00% | 91.67% | 75.00% | 77.08% | | | |
| AHPDegree | | | | | 0.094 | 0.919 | 0.8947 |
| PD | | | | | 0.089 | 0.851 | 0.7788 |

For the overall prediction task, the experimental results demonstrate the effectiveness of the proposed curriculum learning diagnostic prediction framework. In the diagnosis prediction, SciBERT achieved the highest performance, with 97.60% accuracy and 97.14% specificity. BioBERT, on the other hand, achieved the highest F1-score of 97.02% and the highest sensitivity of 97.17%. For AHP type prediction, both SciBERT and PubMedBERT achieved high performance, each reaching 99.78% accuracy, while BioBERT performed slightly lower. The AHP direction task produced slightly reduced scores across models, with PubMedBERT performing best, achieving 96.51% accuracy and 96.04% F1-score. These results indicate greater variability in the descriptive features associated with this label. In the eye classification task, both SciBERT and PubMedBERT achieved 93.46% accuracy, outperforming BioBERT. However, BioBERT achieved the highest performance in the eye misalignment task, with 98.69% accuracy and 98.91% F1-score, reflecting its capability to model fine-grained ocular alignment information within descriptive data.

The regression tasks further highlight the advantages of the curriculum learning strategy under sparse input conditions. PubMedBERT achieved the lowest MAE for both AHP degree (0.249) and PD (0.267), along with the highest $R^2$ values (0.639 and 0.734, respectively), indicating improved generalization in predicting continuous clinical variables. This performance can be attributed to the progressive masking strategy and the domain-aware pretraining of the encoder, which enabled the model to learn effectively from partially observed notes. The structured masking applied at both the token and target levels enabled the model to adapt to increasing levels of input sparsity in a gradual manner. Compared to BioBERT and SciBERT, PubMedBERT demonstrated more stable performance across both classification and regression heads, supporting its selection as the backbone model for the final deployment phase. In addition to MAE and $R^2$, the correlation values further validate the consistency of regression predictions. For AHP degree, BioBERT and PubMedBERT both produced strong correlations of 0.77, while SciBERT achieved the highest at 0.77 as well. In the PD task, BioBERT reached 0.83, PubMedBERT 0.81, and SciBERT 0.80.

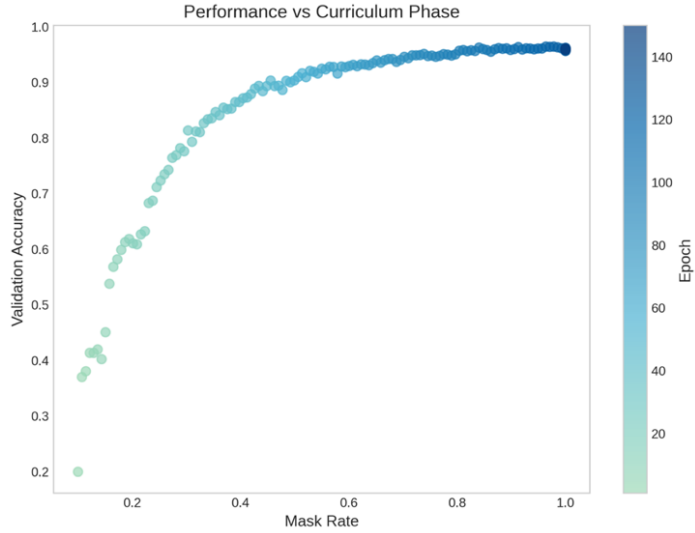

Figure 10: Performance vs Curriculum Phase for PubMed Model

As PubMedBERT achieved a stable performance, more investigation is applied for the results. Figure 10 shows that validation accuracy improved consistently as the curriculum mask rate increased from 0.2 to 1.0, with accuracy rising from approximately 0.20 to 0.95. This trend indicates that the proposed enhancements to the curriculum learning strategy contribute to stable and effective learning under progressively challenging conditions. The absence of performance degradation throughout the progression suggests that the curriculum scheduling successfully enabled the model to generalize from partially observed to fully imputed input scenarios.

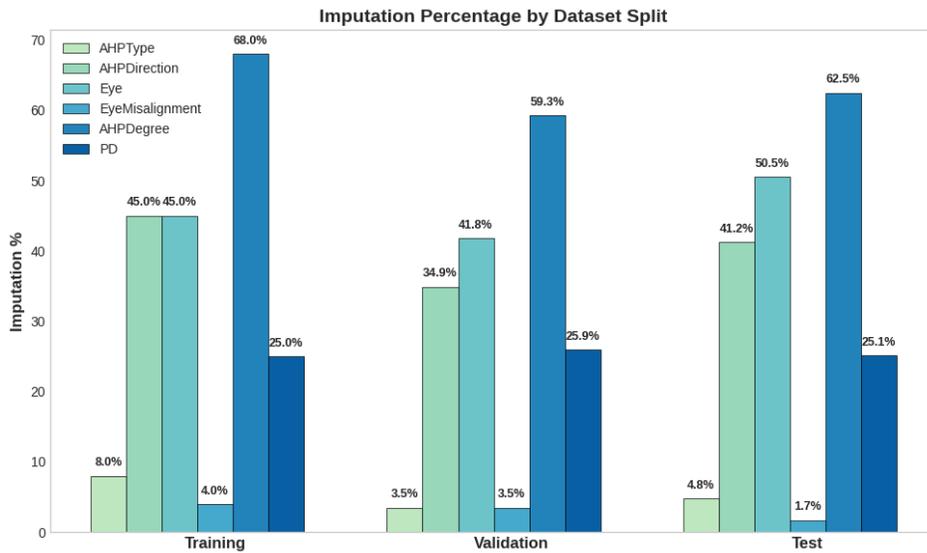

Figure 11: Imputation Percentage Across Clinical Targets and Dataset Split

For the second experiment, model performance was evaluated exclusively on samples with imputed labels to assess generalization under conditions of partial clinical information. Figure 11 illustrates the distribution of missing values across clinical targets in the training, validation, and test splits. Among all targets, the AHP degree consistently exhibited the highest imputation rates, ranging from 59.3% in the validation set to 68.0% in the training set. This was followed by AHP direction, with missing values ranging from 41.8% to 50.5% across the three splits. In contrast, eye misalignment showed minimal missing data, with imputation rates ranging from 1.7% to 4.9%. The AHP type was fully observed in both the validation and test sets, with a modest 8.0% imputation rate, which was present only in the training set.

Despite the varying sparsity, as shown in Table 6, all models achieved high classification accuracy for the AHP type (100%), which can be attributed to the small number of imputed samples and the relatively low complexity of this target. In contrast, performance on targets with higher imputation frequencies showed greater variability. For the AHP direction, PubMedBERT achieved the best performance with an accuracy of 94.71%. Eye classification results were slightly lower across all models, with SciBERT achieving the highest accuracy at 88.36%. Interestingly, for eye misalignment – despite having the fewest imputed samples (n = 60) – predictions were less stable. BioBERT performed best with 87.50% accuracy, while both SciBERT and PubMedBERT achieved 75.00%. These results suggest that frequent imputation does not necessarily lead to improved predictive performance. Instead, the classification challenge appears to be more closely related to the clinical complexity and feature ambiguity of specific targets, such as eye misalignment, rather than to the frequency of imputation.

Regression results further highlight the utility of the curriculum-based architecture under sparse conditions. PubMedBERT achieved the lowest MAE on both AHP degree (0.094) and PD (0.089), along with strong $R^2$ scores of 0.919 and 0.851 and corresponding correlations of 0.89 and 0.78. Despite being the most frequently imputed target, AHP degree predictions obtained robust results, indicating that the progressive masking strategy enabled the model to infer missing continuous values effectively. SciBERT achieved the highest $R^2$ for PD at 0.950 and the strongest correlation for AHP degree at 0.92, though its performance across other tasks was less consistent. BioBERT also produced competitive results, particularly in AHP degree with an MAE of 0.125, an $R^2$ of 0.820, and a correlation of 0.87. Overall, the results demonstrate that the proposed curriculum learning framework supports reliable imputation even in targets with high missingness, with PubMedBERT showing the most consistent performance across classification and regression heads under fully imputed conditions.

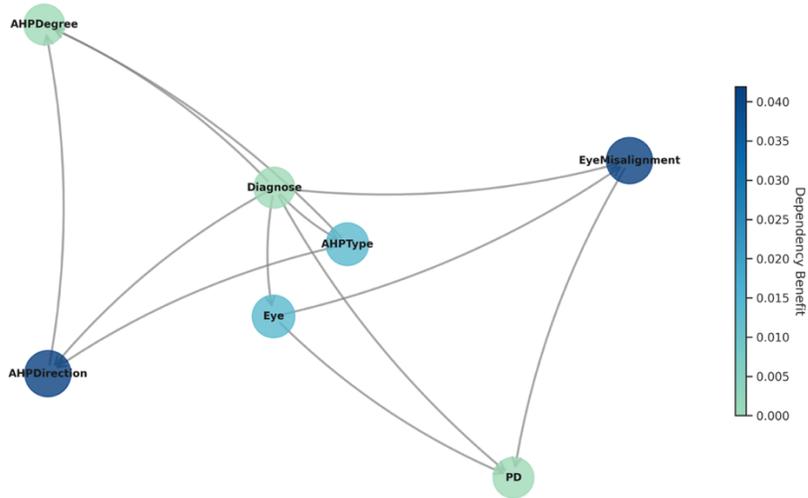

Figure 12: Clinical Dependency Network

Figure 12 illustrates the clinical dependency network, highlighting the information flow across targets within the structured prediction hierarchy. The Diagnose target functions as a complete target and serves as the source node for all downstream predictions. Node color intensity represents the relative benefit obtained from dependency modeling. As shown, targets positioned further along the dependency chain exhibit higher performance gains, with eye misalignment demonstrating the most substantial benefit, followed by AHP direction. The network structure confirms that the model effectively leverages hierarchical relationships, which supports the prediction of targets.

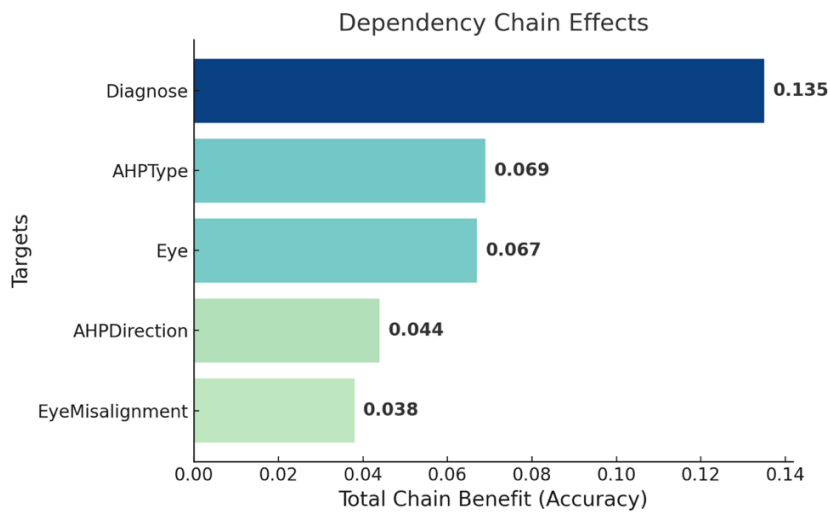

Figure 13: Dependency Chain Effect

Figure 13 quantifies the cumulative effects of dependency chains on prediction performance. The diagnosis target yields the highest total chain benefit (0.135), consistent with its central role in the dependency structure. Intermediate chain effects include eye-to-eye misalignment (0.071) and AHP type to AHP direction (0.060), reflecting clinically plausible relationships where knowledge of primary assessments, such as the affected eye or posture type, enhances the estimation of more detailed clinical features. These results demonstrate that the proposed attention-based dependency layer is capable of capturing and propagating clinically meaningful information across related targets.

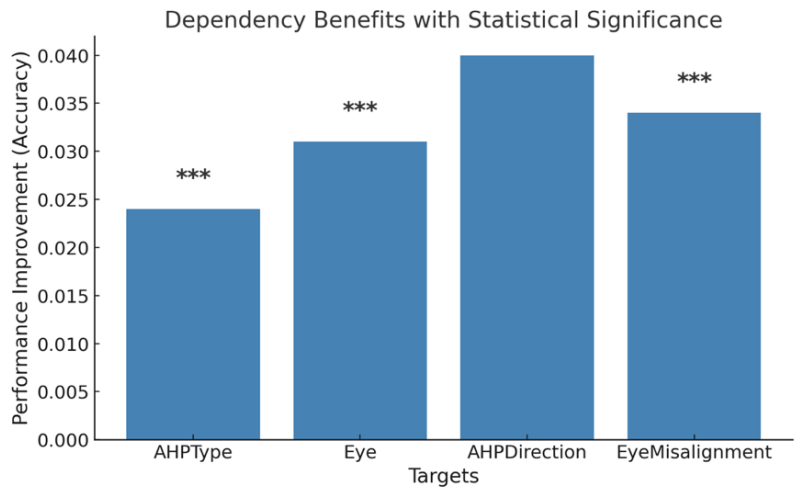

Figure 14: Dependency Benefits with Statistical Significance

Statistical analysis of performance gains, presented in Figure 14, further supports the effectiveness of dependency modeling. All improvements were found to be statistically significant ($p < 0.001$). The greatest gain was observed for AHPDirection (0.040), followed by EyeMisalignment (0.037), Eye (0.032), and AHPType (0.024). The observed gradient in performance improvements aligns with the target hierarchy and reflects increasing reliance on upstream contextual information for targets that are more specific or difficult to observe directly. These findings provide quantitative evidence that modeling clinical target dependencies leads to significant gains in imputation accuracy, particularly for downstream tasks with higher complexity or missing fields.

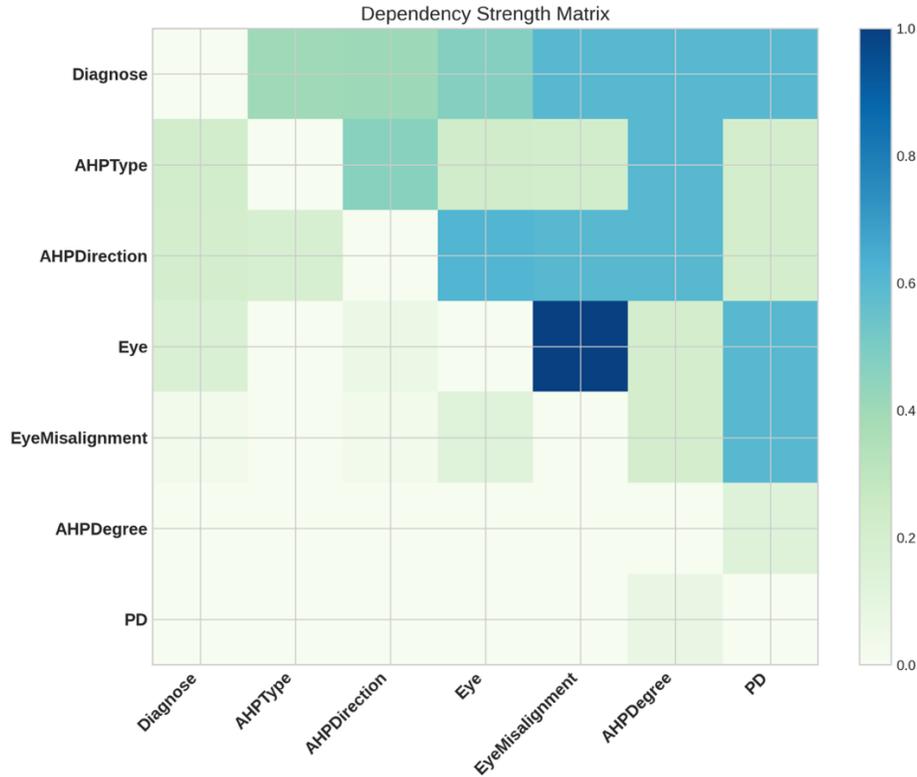

Figure 15: Dependency Strength Matrix

Figure 15 presents the dependency strength matrix, which quantifies the relative contribution of each source target to the prediction of dependent targets. Higher values indicate stronger dependencies, as inferred during the training process. The strongest dependency is observed from eye-to-eye misalignment, indicating that the model effectively captures the clinical relationship, wherein the identification of the affected eye directly informs the classification of the misalignment type. This aligns with established clinical reasoning, as misalignment is often anatomically localized with respect to the affected eye. Moderate improvements are noted from diagnosis to several targets, including AHP direction and AHP degree, reinforcing its role as a fully observed predictor that informs secondary assessments. The relationship between AHP type and AHP direction also demonstrates meaningful strength, consistent with clinical expectations that posture type influences directional alignment patterns. PD exhibits distributed gains from both Eye and eye misalignment, aligning with the understanding that deviation measurements are interpreted in relation to both the involved eye and the nature of the misalignment. A smaller but notable dependency is also observed between PD and AHP degree, where these two continuous measures may share latent associations in AHP severity and ocular deviation.

## Conclusion

This study presents two complementary DL frameworks to address the challenges of automated diagnosis and missing data imputation in ocular-induced AHP. The first, AHP-CADNet, integrates ocular landmarks, head pose features, and structured clinical attributes through a multi-level attention fusion mechanism, achieving robust diagnostic accuracy (96.9%–99.0%) and low error in continuous variable prediction (MAE: 0.103–0.199; $R^2 > 0.93$). The second, a curriculum learning–based imputation framework, which imputes missing clinical data by leveraging both structured variables and unstructured clinical notes, with accuracy (93.46%–99.78%) and statistically significant gains from clinical dependency modeling ($p < 0.001$). These approaches demonstrate competitive performance and hold promise for enhancing the objectivity of DL based ocular-induced AHP diagnosis frameworks.

Despite these promising results, some limitations should be acknowledged. The current study relies on a PoseGaze-AHP dataset with originally 496 cases, which may not fully represent the variability encountered in diverse clinical settings. Differences in EHR structure and documentation across institutions may also affect generalizability of the proposed frameworks. Moreover, the curriculum imputation framework depends on the presence of unstructured clinical notes, which are often inconsistently formatted and vary in linguistic style across clinicians and institutions. Lastly, Finally, the dataset lacks longitudinal data, which restricts the ability to analyze temporal patterns, such as the progression of AHP over time or responses to treatment. These factors may limit this approach when deployed in real-world, heterogeneous clinical environments.

Future research will aim to validate the proposed frameworks using larger, more diverse datasets that more closely reflect real-world EHRs, thereby facilitating clinical applicability. Benchmark datasets will be employed to enable standardized comparisons with existing methodologies. Moreover, the diagnostic framework will be extended to encompass a broader range of AHP etiologies, including neurological and musculoskeletal causes, to enhance its generalizability. Investigating the relationship between AHP and postural adaptations in other body regions may yield deeper insights into compensatory mechanisms. Furthermore, integration into commercial EHR systems and evaluation through prospective clinical trials will be essential to assess workflow integration and patient-level outcomes.

## Acknowledgments

The authors thank the United Arab Emirates University for supporting this work through the UAEU Strategic Research Grant (G00003676) and the Abu Dhabi International Virtual Research Institute for Food Security in the Drylands (G00004017 - VRI-FS 120-21).

## Funding

This research was funded by the United Arab Emirates University Strategic Research Grant G00003676.